\PassOptionsToPackage{numbers,sort&compress}{natbib}
\documentclass{article}
\usepackage{amssymb}

\usepackage[final]{corl_2025} 

\usepackage{xspace}
\usepackage{amsmath}
\usepackage{amsfonts}
\usepackage[capitalise, nameinlink]{cleveref}
\usepackage[dvipsnames]{xcolor}
\usepackage{hyperref}
\hypersetup{
    colorlinks=true,
    linkcolor=orange,
    filecolor=magenta,      
    urlcolor=orange,
    citecolor=orange,
}
\usepackage{booktabs} 
\usepackage{graphicx}
\usepackage{wrapfig}
\usepackage{enumitem}
\usepackage[font=small,labelfont=bf]{caption}
\usepackage{pdfx}
\usepackage{adjustbox}
\usepackage{mathtools}
\usepackage{enumitem}
\usepackage{booktabs}
\usepackage{multirow}
\usepackage{float}

\newcommand{\E}{\mathbb{E}}

\newcommand{\Dt}{\mathcal{D}_\text{t}}
\newcommand{\Dprior}{\mathcal{D}_\text{prior}}
\newcommand{\Dret}{\mathcal{D}_\text{ret}}
\newcommand\Std[1]{{\scriptsize \textcolor{gray}{$\pm$ #1}}}
\newcommand\Out[1]{{\scriptsize \textcolor{gray}{$/#1$}}}

\newcommand{\pt}{p_\text{t}}
\newcommand{\pprior}{p_\text{prior}}
\newcommand{\pret}{p_\text{ret}}

\newcommand{\fullname}[0]{Importance Weighted Retrieval\xspace}
\newcommand{\abv}[0]{IWR\xspace}

\usepackage[subtle]{savetrees}  
\title{Data Retrieval with Importance Weights for Few-Shot Imitation Learning}

\author{
  Amber Xie \\
  Stanford University \\
  \And 
  Rahul Chand \\
  Stanford University
  \And
  Dorsa Sadigh \\
  Stanford University
  \And
  Joey Hejna \\ 
  Stanford University \\
}

\begin{document}
\maketitle


\begin{abstract}
    While large-scale robot datasets have propelled recent progress in imitation learning, learning from smaller task specific datasets remains critical for deployment in new environments and unseen tasks. One such approach to few-shot imitation learning is retrieval-based imitation learning, which extracts relevant samples from large, widely available prior datasets to augment a limited demonstration dataset. To determine the relevant data from prior datasets, retrieval-based approaches most commonly calculate a prior data point's minimum distance to a point in the target dataset in latent space. While retrieval-based methods have shown success using this metric for data selection, we demonstrate its equivalence to the limit of a Gaussian kernel density (KDE) estimate of the target data distribution. This reveals two shortcomings of the retrieval rule used in prior work. First, it relies on high-variance nearest neighbor estimates that are susceptible to noise. Second, it does not account for the distribution of prior data when retrieving data. To address these issues, we introduce \fullname (\abv), which estimates importance weights, or the ratio between the target and prior data distributions for retrieval, using Gaussian KDEs. By considering the probability ratio, \abv seeks to mitigate the bias of previous selection rules, and by using reasonable modeling parameters, \abv effectively smooths estimates using all data points.  Across both simulation environments and real-world evaluations on the Bridge dataset we find that our method, \abv, consistently improves performance of existing retrieval-based methods, despite only requiring minor modifications. \footnote{
Project Website and Code: \url{https://rahulschand.github.io/iwr/}}

\end{abstract}



\section{Introduction}
Data has been integral to the performance of deep learning-based methods across a wide variety of domains \citep{radford2021learning, reed2022a}. Unsurprisingly, the same has found to be true for imitation learning (IL) methods in robotics, for which the most compelling examples often require hundreds to thousands of collected demonstrations in order to learn a single task \citep{zhao2024aloha}. Unfortunately, this makes scaling IL difficult. When trying to learn a new task, one needs to collect a large number of demonstrations to achieve a reasonable level of in-domain success while simultaneously ensuring sufficient diversity for generalization to out-of-distribution scenarios.

One approach for learning from limited demonstrations is \emph{retrieval} from prior datasets. Retrieval-based methods augment small demonstration datasets with relevant samples taken from large, widely available robotics datasets \citep{openx}. This is typically done by learning a representation of all state-action pairs, and ``retrieving'' those from the prior dataset that are most similar to the target demonstration data according to some metric, e.g. distance in latent space \citep{du2023behavior, lin2024flowretrieval,nasiriany2022sailor}. By providing the policy with additional relevant state-action pairs, retrieval reduces the need for collecting additional expert demonstrations for the target task.

Though this has been true in practice, the derivation of retrieval-based methods has largely hailed from intuition. For example, if a data point in the prior dataset is close to that of a target demonstration in latent space, intuitively we can hope that adding it to the training dataset might help the learned policy. 
While this may help justify the performance boost afforded by retrieval-based methods, such design choices are still largely heuristic. In particular, the use of the nearest-neighbor L2 distance metric for scoring prior data is often chosen arbitrarily without principled justification. This begs the question: mathematically, how should we interpret retrieval? Moreover, the possibility remains that a more grounded understanding of retrieval could address the shortcomings of existing heuristic approaches. In this work, we develop such an understanding by taking by a probabilistic perspective through importance sampling, and propose a new method for retrieval. 

At their core, retrieval-based methods aim to leverage a broader data distribution, denoted as $\pprior$, to estimate the loss of a learned policy on the distribution defined by a set of target demonstrations, $\pt$. Usually, to estimate the expectation of a random variable under a target distribution $p$ with samples from an easier-to-sample-form distribution $q$, one would weight samples from $q$ by the ratio of probability densities, $p / q$. Crucially, when dividing by $q$, these ``importance weights'' overcomes the bias introduced by using samples from $q$ instead of $p$. Though retrieval parallels this framework by leveraging samples from $\pprior$ to improve behavior cloning on the task defined by $\pt$, existing retrieval methods can be viewed as only approximating the numerator of this ratio $\pt$, leading to inherent bias. Moreover, the use of the aforementioned nearest-neighbor distance metric is of high variance because of its susceptibility to noise. This leads to two avenues for improvement, which we address through our method, \fullname, or \abv. 

\begin{figure}[t]
\centering
\includegraphics[width=\textwidth]{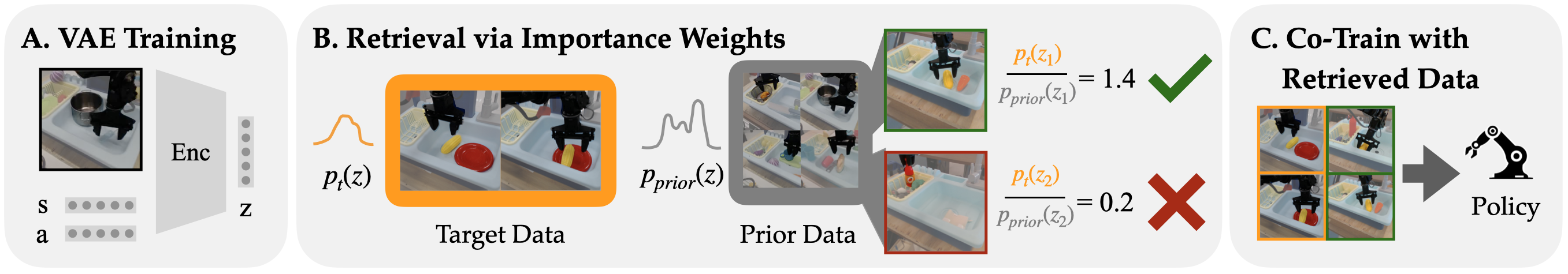}
\caption{\abv consists of three main steps: (A) Learning a latent space to encode state-action pairs, (B) Estimating a probability distribution over the target and prior data, and using importance weights for data retrieval, and (C) Co-training on the target data and retrieved prior data. By augmenting our high-quality but much smaller target dataset with diverse, relevant prior samples, we learn more robust and performant policies.}
\label{fig:method}
\vspace{-0.25in}
\end{figure}

\abv simultaneously addresses both the bias and high-variance of prior L2-distance based retrieval methods by applying Gaussian kernel density estimation (KDE) to estimate the full importance weights $\pt / \pprior$. Using Gaussian KDEs produces smoother estimates by considering all data points within a dataset instead of just the nearest-neighbor (see \cref{fig:is_l2}). Then, by using KDEs to approximate both $\pt$ and $\pprior$, \abv uses importance weights to retrieve data, seeking a more accurate approximation of the expectation under $\pt$ (\cref{fig:method}). 

Practically, we find that these choices allow \abv to retrieve higher quality data, in terms of relevancy to the target task and diversity across task phases. Moreover, these benefits are not limited to a single retrieval-based IL method -- we find that \abv consistently improves the quality of retrieved data when applied to a number of different prior works by simply replacing nearest-neighbor distance queries with estimated importance weights. For example, on the LIBERO benchmark, using \abv increases average success rates by 5.8\% on top of SAILOR \citep{nasiriany2022sailor}, 4.4\% on top of Flow Retrieval \citep{lin2024flowretrieval}, and 5.8\% on top of Behavior Retrieval \citep{du2023behavior}. On real-world tasks with the Bridge V2 Dataset \citep{bridge}, we find the performance improvement afforded by \abv to be more significant, increasing success rate by 30\% on average, in comparison to Behavior Retrieval.

\section{Related Work}

\textbf{Retrieval} Retrieval enables few-shot imitation learning by augmenting a target dataset, consisting of a handful of demonstrations for a task, with additional prior data from pre-existing datasets such as DROID~\citep{khazatsky2024droid}, OpenX~\citep{openx}, and Bridge~\citep{bridge}. Typically, the retrieved prior data is used to augment the dataset for imitation learning~\citep{du2023behavior,lin2024flowretrieval,memmel2025strap}, with the focus of these works primarily about retrieval, not algorithmic improvements~\citep{nasiriany2022sailor,lin2024flowretrieval}. BehaviorRetrieval~\citep{du2023behavior} and FlowRetrieval~\citep{lin2024flowretrieval} both learn a latent embedding space of state-action pairs and optical flow respectively, and they retrieve prior data points that are closest to the latent embeddings of the target data points. SAILOR~\citep{nasiriany2022sailor} learns latent embeddings through skill-based representation learning, and they also retrieve data points that are closest to the target data in latent space. STRAP~\citep{memmel2025strap} uses foundation model latent embeddings and retrieves sub-trajectories based on dynamic time warping and proximity in latent space. While most of these works focus on altering latent representations, our work instead focuses on improving the Euclidean distance metric typically used for retrieval. 

\textbf{Importance Sampling}
Outside of few-shot imitation learning, importance sampling, from which \abv is motivated, has been used for data selection in a variety of domains. Works in imitation learning use importance weights to discard sub-optimal transitions \cite{yue2024leverage} or for off-policy evaluation \citep{hanna2019importance}. In reinforcement learning, importance weights have been used for prioritized sampling \citep{sinha2022experience}. Perhaps most similar to the retrieval problem, importance weights have been used to select relevant documents for training language models \citep{xie2023data}. While more complicated techniques for estimating importance weights have been developed, from telescoping classifiers \citep{BENNETT1976245, NEURIPS2020_33d3b157} to generative models \citep{choi2022density}, we find simple Gaussian KDEs to be effective.

\textbf{Data for Imitation Learning}
The paradigm of co-training policies with large amounts of additional, diverse \citep{openx, khazatsky2024droid} or even simulated \citep{maddukuri2025simandrealcotrainingsimplerecipe} data has proven effective in imitation learning. While these works take a blanket approach, retrieval seeks to identify the most relevant data for a particular tasks. Other works have sought to identify relevant data, but with different goals in mind. \citet{hejna2024remix} identify group weights in large robot datasets to accelerate behavior cloning. Others filter data based on various quality metrics \citep{belkhale2024data} using mutual information \citep{hejna2025robot}, preferences \citep{kuhar2023learning}, or demonstrator expertise \citep{beliaev2022imitation}. Instead of selecting data, other works opt to generate similar data leveraging simulation \citep{ha2023scalingup, mandlekar2023mimicgen}.

\section{Preliminaries}

\subsection{Problem Setup}
The objective of imitation learning (IL) methods is to learn a policy $\pi(a|s)$, which mimics the expert behavior within an environment with states $s$ and actions $a$. Typically, the policy is learned using a dataset $\mathcal{D}$ of expert trajectories $\tau = \{s_0, a_0, \dots, s_T, a_T\}$, where $s_t, a_t$ correspond to states and actions at timestep $t$, and $T$ is the length of the trajectory. While IL has shown success in simulated and real tasks, collecting the requisite expert demonstrations for IL is expensive and has to be repeated for each new task one wishes to learn. 

Retrieval-based methods have sought to address this shortcoming by leveraging existing prior data when learning a new task. Specifically, we consider a few-shot setting in which we only have a handful of expert demonstrations for a new target task, our target dataset $\Dt$. Retrieval-based methods assume access to a much larger prior dataset $\Dprior$, consisting of more diverse tasks and scenes. To decrease the number of demonstrations needed in $\Dt$, retrieval-based methods carefully select data from $\Dprior$, which is then used to co-train the policy. We denote this dataset of retrieved state-action pairs as $\Dret \subset \Dprior$. Mathematically, this leads to the following weighted behavior cloning objective for a parameterized policy $\pi_\theta$: \looseness=-1
\begin{equation}
    \max_\theta \quad \alpha \frac{1}{|\Dt|} \sum_{(s,a) \in \Dt} \log \pi_\theta(a|s) + (1 - \alpha) \frac{1}{|\Dret|} \sum_{(s,a) \in \Dret} \log \pi_\theta(a|s)
    \label{eq:retrieval_bc}
\end{equation}
where $\alpha$ (typically 0.5) is the weighting coefficient between the target and retrieved data. By training on additional data, retrieval-based methods aim to increase the robustness of the learned policy. 

Though retrieval methods often differ in the specific representations they learn, most share a common selection mechanism: L2 distance between embedding representations $z$ of the target and prior data. To learn latent embeddings, BehaviorRetrieval~\citep{du2023behavior} trains a variational autoencoder (VAE) over state-action pairs $(s_t, a_t)$, FlowRetrieval~\citep{lin2024flowretrieval} learns a VAE over the optical flow for each state $s_t$, and SAILOR~\citep{nasiriany2022sailor} encodes a sub-trajectories $(s_t, a_t, \dots, s_{t+k}, a_{t+k})$ using skill-based representation learning. Despite differences in how these latent spaces are learned, all of these methods select data based on the L2 distance. Denoting the learned encoders as $f_\phi$ and the representations they produce as $z = f_\phi(s, a)$ the selection rule can be written as: \looseness=-1

\begin{equation}
    \Dret \coloneqq \left\{ (s,a) \in \Dprior \middle|  {\textstyle \min_{(s', a') \in \Dt}} \left\lVert 
f_\phi(s, a) - f_\phi(s', a') \right\rVert_2^2 < \zeta \right\}
\label{eq:retrieval_rule}
\end{equation}
where $\zeta$ is the retrieval threshold, often chosen such that $\Dret$ is small percentage of $\Dprior$. This intuitively makes sense, as data with representations that are close to those of $\Dt$ are likely useful for training. In the next section, we re-examine retrieval through a probabilistic lens. 

\subsection{From samples to densities}
\label{sec:probabilistic_retrieval}
To characterize retrieval probabilistically, we define marginal state-action distributions $\pt$, $\pprior$, and $\pret$, from which we assume $\Dt$, $\Dprior$, and $\Dret$ are sampled. Then, the previous IL objective (\cref{eq:retrieval_bc}) becomes
\begin{equation}
    \max_\theta \quad \alpha \E_{(s,a) \sim \pt}\left[ \log \pi_\theta (a|s) \right] + (1 - \alpha) \E_{(s,a) \sim \pret}\left[ \log \pi_\theta (a|s) \right].
\label{eq:probabilistic_retrieval}
\end{equation}
However, for a given target task, we are primarily interested in maximizing the policy likelihood under the target task distribution, i.e. $\E_{(s,a) \sim \pt}\left[ \log \pi_\theta (a|s) \right]$. In order to maximize this policy likelihood when optimizing \cref{eq:probabilistic_retrieval}, we would like retrieval-based methods to align the distribution of retrieved data such that it is equivalent to that of the target task, i.e. $\pt \approx \pret$. Then, \cref{eq:probabilistic_retrieval} would amount to behavior cloning on the target task distribution, which is our desired objective, while using the additional samples from $\Dret$, which we would expect to improve performance. 

Examining the selection rule \cref{eq:retrieval_rule} used in prior work, we find that it considers the minimum squared distance to a data point in $\Dt$, corresponding to an approximation of $\pt$, as it does not leverage samples from $\Dprior$ (we formalize this connection in in \cref{sec:kde}). First, we note that this approximation of $\pt$ is imprecise as it only uses the nearest neighbor in $\Dt$, resulting in high variance and susceptibility to noise. Second, even if \cref{eq:retrieval_rule} were able to perfectly recover $\pt$, it still only uses estimates of $\pt$ to retrieve from $\Dprior$. The resulting retrieved distribution $\pret$ is thus closer to the product of densities $\pt \cdot \pprior$ than $\pt$, as the prior data is first sampled according to $\pprior$, and then retrieved according to a condition of $\pt$. We address these limitations through \abv, ensuring that we can retrieve a more accurate approximation of the desirable distribution.

\section{\fullname}

In this section, we address the shortcomings of the standard retrieval selection rule by introducing our method \fullname (\abv). Similar to other retrieval methods, we assume access to an embedding function $f_\phi$, typically taken from a VAE. Given the resulting embeddings from $f_\phi$, we first discuss how we can better model the distributions used in retrieval with Gaussian KDEs. Second, we discuss how modeling the prior distribution $\pprior$, which we use to compute importance weights $\pt / \pprior$, allows us to retrieve data from the desired distribution $\pt$. Our approach can be applied in conjunction with a broad set of retrieval-based methods to improve performance. \looseness=-1

\subsection{Improved Density Modeling}
\label{sec:kde}
\begin{wrapfigure}[21]{R}{0.35\textwidth}
    \vspace{-1.5em}
    \centering
    \includegraphics[width=\linewidth]{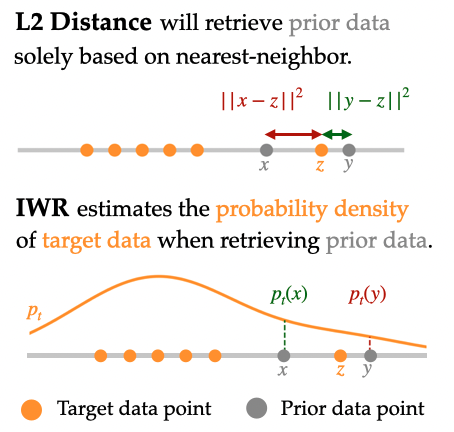}
    \caption{In this toy example, using L2 distance in latent space leads to the left point \textcolor{BrickRed}{discarded}, and the right point \textcolor{OliveGreen}{retrieved}. However, when using \abv to estimate the probability density of the target data, the left point is \textcolor{OliveGreen}{retrieved}. This is because \abv has a smoothing effect and uses many target points for retrieval. In this example, we may expect the left point to be  relevant, as it is close to many target points, as \abv correctly determines.}
    \label{fig:is_l2}
\end{wrapfigure}

As discussed in \cref{sec:probabilistic_retrieval}, standard retrieval methods select data points from $\Dprior$ that minimize the L2 distance from their nearest neighbors in $\Dt$, which may suffer from high variance. To address this, we use lower variance estimates of the probability density function (pdf), which considers all data points (\cref{fig:is_l2}). 

The condition from \cref{eq:retrieval_rule} can equivalently be written as follows using a max instead of a min: $\max_{(s', a') \in \Dt} -\left\lVert 
f_\phi(s, a) - f_\phi(s', a') \right\rVert_2^2 > -\zeta$.
To smooth this retrieval rule, we replace the hard maximum with a log-sum-exp parameterized by temperature $h$, resulting in a soft approximation that aggregates contributions from all data points. The retrieved dataset then becomes:
\begin{equation}
    \Dret \coloneqq \left\{ (s,a) \in \Dprior \middle|  \tfrac{1}{h^2} \log {\textstyle \sum_{(s', a') \in \Dt}}  \exp \left(-\left\lVert  f_\phi(s, a) - f_\phi(s', a') \right\rVert_2^2 \middle/ h^2 \right) > -\zeta \right\}.
\label{eq:lse}
\end{equation}
Here, the exponential term within the sum is proportional to the multivariate Gaussian pdf $\mathcal{N}$, with mean $f_\phi(s', a')$ and covariance matrix $h^2 I$, evaluated at $f_\phi(s, a)$. 
This implies that the sum over all such Gaussians -- each centered at each data-point in $\Dt$ -- is proportional to a Gaussian kernel density estimate (KDE) of $\Dt$, assuming isotropic covariance $I$ and bandwidth $h$. In the limit as the bandwidth $h \rightarrow 0$, the KDE becomes sharply peaked at each data point in $\mathcal{D}$, and the density at a point is dominated by its nearest neighbor -- recovering the original retrieval rule from \cref{eq:probabilistic_retrieval}.
Under this view, the original retrieval rule can be interpreted as a limiting case of a KDE estimate of $\pt$ implying that prior retrieval methods implicitly rely on overly restrictive modeling assumptions. 

Instead, we directly model distributions with Gaussian KDEs using well-calibrated parameters that smooth across data points to obtain lower variance estimates (See \cref{fig:is_l2}). Specifically, we employ bandwidths $h$ set to multiplicative factor of Scott's rule \citep{scott2015multivariate} and use the sample covariance matrix $\Sigma$, giving us \looseness=-1

\begin{equation}
\label{eq:kde}
    p^\text{KDE}(z) = \tfrac{1}{|\mathcal{D}|} {\textstyle \sum_{z' \in f_\phi(\mathcal{D})}}  \left((2\pi)^{d} |h^2 \Sigma| \right)^{-1/2} \exp\left\{-\tfrac{1}{2} (z - z')^\top (h^2 \Sigma)^{-1} (z - z') \right\}
\end{equation}
where $z$ and $z'$ are the representations of state-action pairs from $f_\phi$ and $f_\phi(\mathcal{D})$ denotes an encoded dataset. In comparison to the original retrieval rule, using \cref{eq:kde} to model $\pt$ prefers retrieving data points near multiple targets and better handles dependencies among features via $\Sigma$.

\subsection{Importance Weighting}
Our ultimate goal in retrieval is to estimate the expectation of the loss under the target distribution $\pt$ using samples from the prior distribution $\pprior$. This bears a striking resemblance to importance sampling. Instead of selecting data according to $\pt$, we select data according to the importance weight $\pt / \pprior$ as
\begin{equation}
    \E_{\pprior} \left[ \pt / \pprior \log \pi (a|s) \right] = \E_{\pt} \left[\log \pi (a|s) \right]
    \label{eq:our_method}
\end{equation}
which ensures that the expectation under samples from $\pprior$ is the same as that of $\pt$. Practically, we can use the aforementioned KDE estimators to fit importance weights as $\pt^\text{KDE} / \pprior^\text{KDE}$ following \cref{eq:kde}. Doing so overcomes the bias in the retrieval distribution introduced by the prior dataset $\Dt$. 

Then, given a particular threshold we select data points from $\Dprior$ which have the highest estimated importance weights $\pt^\text{KDE} / \pprior^\text{KDE}$. Though consistent with prior retrieval works, it is not an unbiased estimate. Alternatively, we could obtain an unbiased estimate by following an importance resampling procedure \citep{gelman2004applied}, where $K$ data points are sampled from $\Dprior$ based on estimated importance weights. However, such an approach could be higher variance due to the nature of importance sampling procedures, and we thus follow the simple thresholding procedure that easily integrates with prior methods. 

\subsection{Putting It All Together}
Beginning with a handful of demos in $\Dt$, the final recipe for \abv involves the following steps:

\noindent \textbf{1. Representation Learning.} First, we train a model $f_\phi$ to produce low-dimensional representations $z$ of state-action pairs or sequences. Most prior works in retrieval address this step. For example, SAILOR uses ``skill'' representation learning while Flow Retrieval uses VAEs on learned from visual flow. \abv is compatible with all types of representation learning, so long as the learned latent dimension is sufficiently small for a Gaussian KDE. In this manner, \abv can be combined with several prior works in retrieval. 

\noindent \textbf{2. Importance Weight Estimation.} Following \cref{eq:kde}, we fit Gaussian KDEs to embedding representations of $\Dt$ and $\Dprior$ to estimate both $\pt \approx \pt^\text{KDE}$ and  $\pprior \approx \pprior^\text{KDE}$. Then, we query the KDEs at $z$ for embedded state-action chunks or sequences in $\Dprior$ to estimate their importance weights $\pt^\text{KDE}/\pprior^\text{KDE}$. In practice, $\Dprior$ is often too big to fit with a single KDE, so we estimate $\pprior^\text{KDE}$ using random batches from $\Dprior$.  \looseness=-1

\noindent \textbf{3. Data Retrieval.} We then select data with the highest importance weights to train on, following a similar rule to that of \cref{eq:retrieval_rule} except with importance weights; e.g. $\pt^\text{KDE}/\pprior^\text{KDE} > \eta$. Similar to prior work, the threshold $\eta$ is determined experimentally or by examining the distribution of scores. \abv is still a biased estimate, as we do not perform importance sampling, but only retrieve based on importance weights. \looseness=-1

\noindent \textbf{4. Policy Learning.} Finally, we co-train our policy with the retrieved data following \cref{eq:retrieval_bc}.

In comparison to contemporary methods, \abv has minimal overhead as it only modifies the latter steps in retrieval for importance weight estimation.

\section{Experiments}
In this section we answer the following questions: 1) How much does \abv improve performance? 2) Is \abv broadly applicable to all retrieval methods? 3) What contributes to \abv's performance?

\subsection{Experimental Setup}

\begin{wrapfigure}[15]{R}{0.35\textwidth}
    \vspace{-2.5em}
    \centering
    \includegraphics[width=\linewidth]{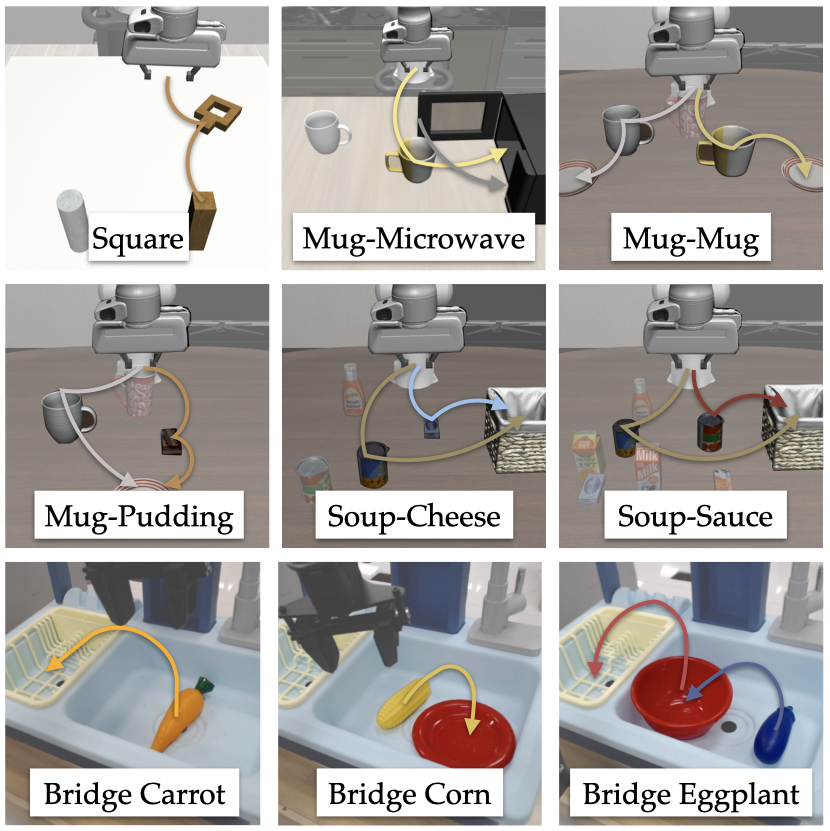}
    \caption{We evaluate on simulated environments: Robomimic Square, a suite of 5 LIBERO-10 tasks, which each task consisting of two subtasks. For our real experiments, we consider 3 Bridge tasks, with Eggplant being a long-horizon task.}
    \label{fig:tasks}
\end{wrapfigure}

\textbf{Simulated Tasks.} We evaluate on two simulated domains used in prior retrieval work: Robomimic Square~\citep{robomimic2021} and LIBERO~\citep{liu2023libero}. 

\begin{itemize}[leftmargin=1em, noitemsep, topsep=1pt]
\item \textbf{Robomimic Square:} We select the Square Assembly task from Robomimic, a popular imitation learning benchmark. We use the same datasets as Behavior Retrieval \citep{du2023behavior}, where $\Dt$ consists of 10 demonstrations lifting the square nut into the goal peg, and $\Dprior$ consists of 400 trajectories, with 200 placing the square nut on the goal peg, and 200 adversarial episodes with the wrong peg. Similar to prior work \citep{du2023behavior,lin2024flowretrieval}, we retrieve 30\% of $\Dprior$.

\item \textbf{LIBERO:} We use the LIBERO benchmark~\citep{liu2023libero}, and select the 5 tasks from LIBERO-10 with the lowest non-trivial success from~\citep{memmel2025strap}, where each $\Dt$ consists of 5 demos. Following ~\citep{memmel2025strap}, we use LIBERO-90 as $\Dprior$ and condition our policy on one-hot task vectors. For each task we retrieve 2.5\% of the prior data.
\end{itemize}

\textbf{Real World Tasks.} We further instantiate experiments in the real world using the Bridge setup~\citep{bridge}. We include 3 Bridge tasks: \textit{Corn} with 5 demos, where the robot is tasked with moving a corn onto a plate; \textit{Carrot} with 10 demos, where the robot moves a carrot into the sink, and \textit{Eggplant} with 20 demos; a long-horizon task consisting of grasping the eggplant, placing it into the bowl, and transferring the bowl with the eggplant into the dish rack. These three tasks are unseen in the Bridge-V2 dataset. However, the prior dataset consists of similar tasks performed in many toy sinks, including the specific toy sink for our tasks, and thus we may expect transfer from this prior dataset.

Our $\Dt$ consists of VR-teleoperated demonstrations on a WidowX, and our $\Dprior$ consists of a subset of Bridgev2 Dataset~\citep{bridge}, where we take trajectories with keyword sink, leading to 130k transitions.

\textbf{Baselines} We compare \abv to popular retrieval baselines:

\begin{itemize}[leftmargin=1em, noitemsep, topsep=0pt]
    \item \textbf{Behavior Cloning (BC)} is only trained to imitate $\Dt$. 
    \item \textbf{Behavior Retrieval (BR)} learns a VAE over state-action pairs, and it retrieves data from $\Dprior$ based on L2 distance in the latent space. 
    \item \textbf{Flow Retrieval (FR)} learns a VAE over optical flow between frames and actions. Similar to BR, it retrieves data via L2 distance in latent space.
    \item \textbf{SAILOR (SR)} learns a skill-based latent space by compressing state-action chunks. SAILOR retrieves data via L2 distance in latent space.
\end{itemize}
We evaluate the performance of different methods by training Diffusion Policy \citep{chi2023diffusion} on the retrieved and target data following \cref{eq:retrieval_bc}. By default, we use the representations from BR for \abv, as we found them to perform best overall. More details can be found in the Appendix.

\begin{table}[t]
\caption{We evaluate across a suite of \textit{simulated} (Square, Mug-Microwave, .., Soup-Sauce) and \textit{real} (Corn, .., Eggplant) tasks. We find that \abv consistently outperforms previous retrieval baselines. We report success rates in \% over 3 seeds for simulated tasks. For real-world tasks, we report success rate over 20 trials. For the long-horizon Eggplant task, we also record Partial Success (PS) for completing subtasks. We bold the best-performing method.
}
\label{tab:sim}
\begin{adjustbox}{width=1\textwidth,center=\textwidth}
\centering
\begin{tabular}{l|c|ccccc|cccc}
\toprule
\textbf{Method} & \textbf{Square} & \textbf{Mug-} & \textbf{Mug-} & \textbf{Mug-} & \textbf{Soup-} & \textbf{Soup-} & \textbf{Corn} & \textbf{Carrot} & \textbf{Eggpl.} & \textbf{Eggpl.} \\
& & \textbf{Microwave} & \textbf{Mug} & \textbf{Pudding} & \textbf{Cheese} & \textbf{Sauce} & & &  \textbf{Partial} & \textbf{Full} \\
\midrule
BC & 1 \Std{0.9} & 72 \Std{0.9} & 54 \Std{3.3} & 21 \Std{2.4} & 58 \Std{6.8} & 32 \Std{2.5} & 4\Out{20} & 2\Out{20} & 9\Out{20} & 2\Out{20} \\
BR & 69 \Std{5.0} & \textbf{81} \Std{0.5} & 81 \Std{2.4} & 33 \Std{3.3} & \textbf{83} \Std{4.5} & 43 \Std{2.7} & 2\Out{20} & 8\Out{20} & 8\Out{20} & 3\Out{20} \\
SR & 40 \Std{4.9} & 67 \Std{2.4} & 67 \Std{4.7} & 14 \Std{0.9} & 76 \Std{4.3} & 51 \Std{2.2} & \textbf{12\Out{20}} & 3\Out{20} & 10\Out{20} & 2\Out{20} \\
FR & 79 \Std{5.0} & 79 \Std{2.2}  & 59 \Std{4.3}  & 17 \Std{2.9} & 37 \Std{3.8} & 45 \Std{5.5} & 2\Out{20} & 3\Out{20} & 6\Out{20} & 0\Out{20} \\
\abv & \textbf{84} \Std{2.8}  & \textbf{81} \Std{3.6} & \textbf{87} \Std{2.0} & \textbf{45} \Std{1.4} & \textbf{83} \Std{3.3} & \textbf{54} \Std{5.7} & 9\Out{20} & \textbf{14\Out{20}} & \textbf{20\Out{20}} & \textbf{11\Out{20}} \\     
\bottomrule
\end{tabular}
\end{adjustbox}
\vspace{-0.25in}
\end{table}

\begin{figure}
    \centering
    \includegraphics[width=0.49\linewidth]{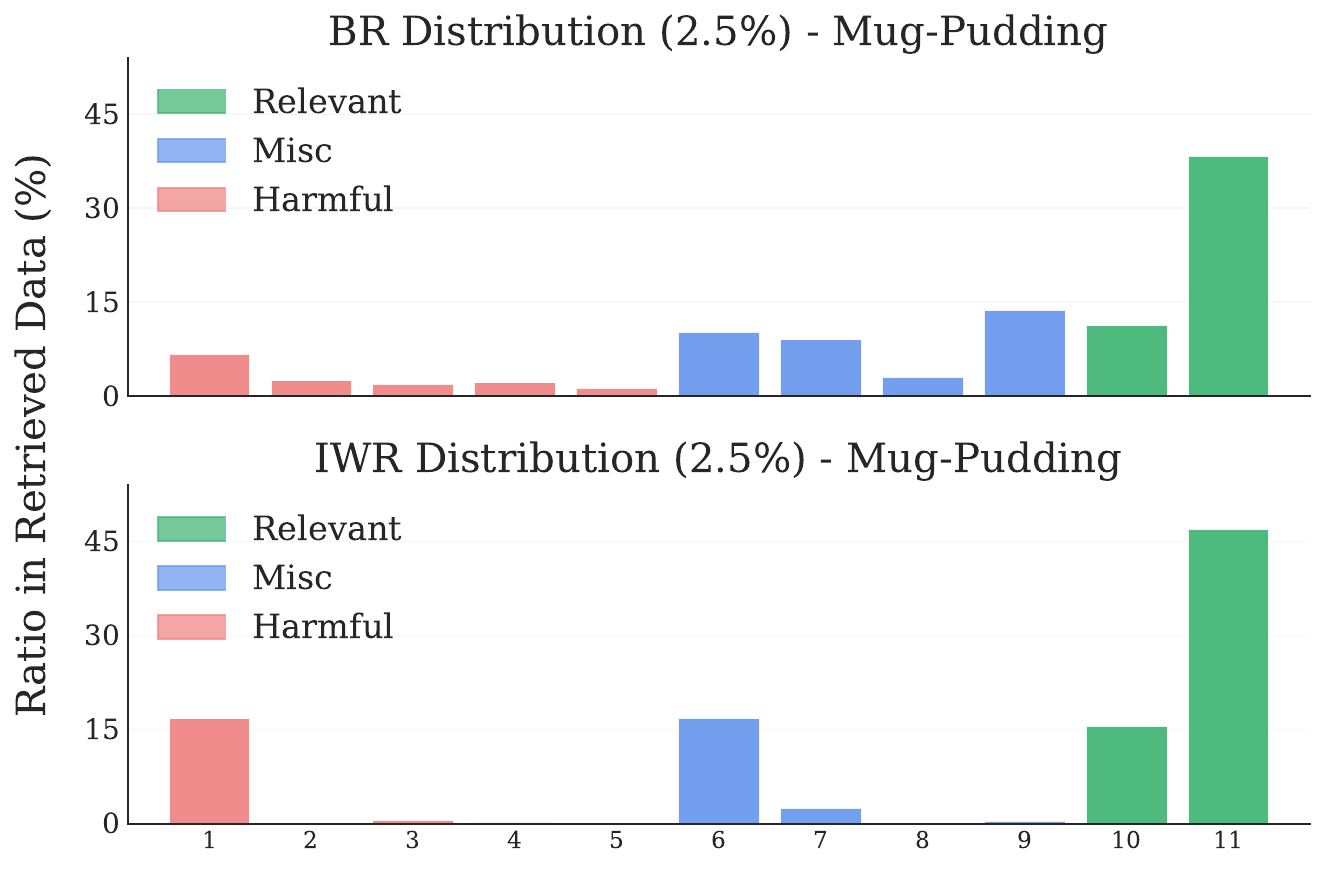}
    \includegraphics[width=0.49\linewidth]{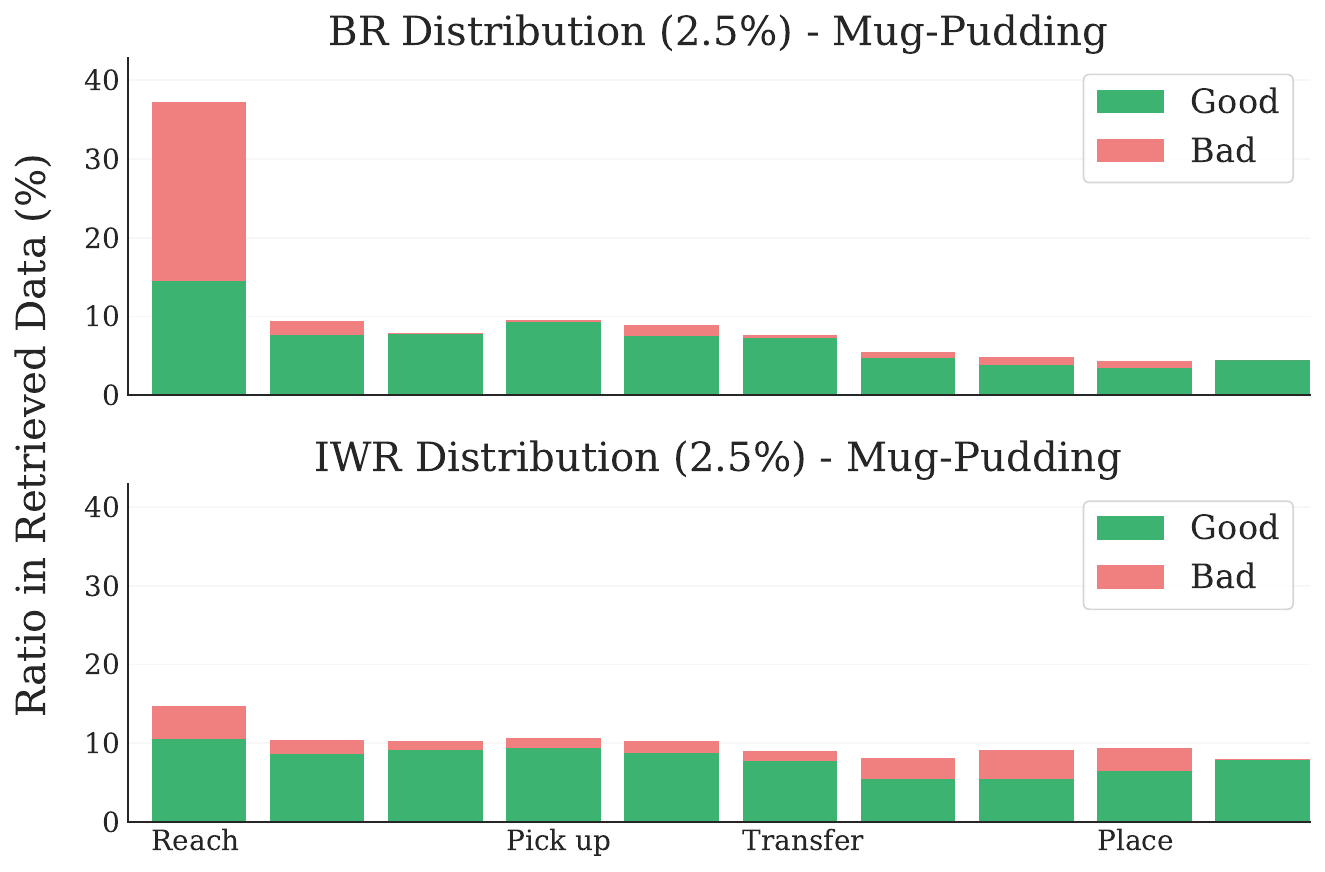}

    \includegraphics[width=\linewidth]{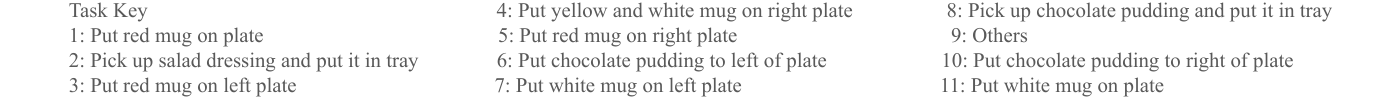}
    \caption{Difference in retrieval distributions between BR and \abv for the Mug-Pudding task in terms of both tasks (left) and timesteps (right). (Left) Prior tasks which form exact sub-tasks of the target are marked as Relevant, tasks with at least one common object with target as Misc, and the rest as Harmful. (Right) Demonstrations are divided into 10 bins. Green represent samples from either relevant or temporally-appropriate portions of partially-relevant Misc tasks. More details can be found in the Appendix.}
    \label{fig:tasks-ret}
    \vspace{-0.25in}
\end{figure}

\subsection{How much does \abv improve performance?}

\textbf{Simulated Experiments.} The left half of~\cref{tab:sim} provides results for simulated tasks. In Robomimic Square, retrieval is crucial, as training only on $\Dt$ leads to extremely poor performance. Here, where retrieving incorrect data is detrimental, \abv consistently outperforms other methods, likely due to its lower variances estimates of $\pt$. Only FR exhibits similar performance, likely because the prior data includes only two motions (peg to left or right) which is readily captured by visual flow. In LIBERO, where $\Dprior$ is significantly larger and more diverse, SR and FR perform considerably worse, often under-performing BC.
For LIBERO, BR faces two challenges: (1) object similarity across tasks results in retrieval of irrelevant demonstrations, and (2) tasks often share similar starting configurations, which bias retrieval towards initial samples instead of more informative later-stage actions. Though these issues are correlated, even if we only retrieve from relevant tasks, the bias towards starting samples can still compromise performance.

\abv addresses these limitations by upweighting samples containing underrepresented objects or occurring later in the demonstrations. Consider the Mug-Pudding task, which requires placing a white mug on a plate and a pudding on the left of the plate. In this case, ~\cref{fig:tasks-ret} shows that BR retrieves irrelevant tasks containing either ``chocolate pudding'' or ``white mug'' and also disproportionately samples from the initial phase ($\sim$40\%). \abv corrects both issues, retrieving a higher percentage of relevant tasks and a more balanced distribution across timesteps as importance weights correct for the bias in $\Dprior$.

On the other hand, for Mug-Microwave (place mug inside microwave), BR and \abv perform similarly because the critical subtask of ``inserting into microwave'' is absent from all priors, resulting in a consistent failure mode where objects collide with the microwave. This failure mode cannot be overcome with better or more sampling. In contrast, for Soup-Cheese, BR already achieves high performance due to the task's simplicity and distinctive priors - one of the task's component involves a cheese box that is visually different from other objects in similar task setting (cans, ketchup). With both BR and IWR retrieving over 50\% from directly relevant tasks, \abv's improved retrieval offers minimal additional benefit.

\textbf{Real World Experiments.} We evaluate \abv and baselines on a real-world Bridge sink environment, with performance on the three tasks reported in \cref{tab:sim}. For all of these tasks, using retrieved prior data improves performance. Qualitatively, the BC policies often early grasp or miss the object, which additional pick-and-place retrieved data helps mitigate. \abv consistently leads to the largest improvements, especially for the long-horizon Eggplant task, where \abv is able to achieve partial success on 100\% of rollouts, compared to the best performing retrieval method, which only achieves partial success on 50\% of rollouts. For this longer horizon task, the effect of \abv's data retrieval leads to a drastic improvement, as additional data may alleviate issues with accumulating errors across subtasks. 

Among other retrieval methods, FR performs poorly across most tasks, whereas SR is particularly successful for Corn. We hypothesize that for tasks in more cluttered scenes, such as Corn, FR's embeddings are not precise enough, because the optical flow may be noisy. SR's performance in Real may be because Bridge tasks have lower-frequency control, so its skill-based, chunking representations consist of more cohesive motion than in simulated tasks.

\begin{table}[]
\caption{We ablate existing retrieval embeddings with importance weights for retrieval (\abv).}
\label{tab:l2_is}
\begin{adjustbox}{width=1\textwidth,center=\textwidth} 
\begin{tabular}{l|c|ccccc|cccc}
\toprule
\textbf{Method} & \textbf{Square} & \textbf{Mug-} & \textbf{Mug-} & \textbf{Mug-} & \textbf{Soup-} & \textbf{Soup-} & \textbf{Corn} & \textbf{Carrot} & \textbf{Eggpl.} & \textbf{Eggpl.} \\
& & \textbf{Micro.} & \textbf{Mug} & \textbf{Pudding} & \textbf{Cheese} & \textbf{Sauce} & & & \textbf{Partial} &  \textbf{Full} \\
\midrule
SR & 40 \Std{4.9} & 67 \Std{2.4} & 67 \Std{4.7} & 14 \Std{0.9} & 76 \Std{4.3} & 51 \Std{2.2} & 12\Out{20} & 3\Out{20} & 10\Out{20} & 2\Out{20} \\
SR-\abv & 57 \Std{9.4} & 81 \Std{3.0} & 70 \Std{3.4} & 20 \Std{0.9} & 85 \Std{1.4} & 48 \Std{3.4} & 9\Out{20} & 13\Out{20} & 16\Out{20} & 1\Out{20} \\
\midrule
FR & 79 \Std{5.0} & 79 \Std{2.2} & 59 \Std{4.3} & 17 \Std{2.9} & 37 \Std{3.8} & 45 \Std{5.5} & 2\Out{20} & 3\Out{20} & 6\Out{20} & 0\Out{20} \\
FR-\abv & 67 \Std{1.9} & 81 \Std{4.8} &  69 \Std{1.1} & 18 \Std{0.9} & 49 \Std{3.6} & 42 \Std{3.3} & 3\Out{20} & 11\Out{20} & 8\Out{20} & 0\Out{20} \\
\bottomrule
\end{tabular}
\end{adjustbox}
\vspace{-0.2in}
\end{table}

\subsection{\abv with Other Retrieval Methods}
Though we use the latent space from BR for \abv in ~\cref{tab:sim}, \abv can be combined with other learned representations such as those from FR and SR with minimal modification; we simply estimate importance weights using KDE's for retrieval instead of using L2 distance. In \cref{tab:l2_is}, we combine \abv with SR and FR for simulated and real tasks. We note that these latent spaces are all learned through a VAE objective, which encourages smoothness and is common for retrieval. We find \abv does not lead to improvements for a non-smooth latent space, detailed in the ~\cref{tab:byol} in the Appendix.

Across simulated tasks, we generally find that \abv augmented retrieval leads to stronger policy performance. In real, adding \abv to FR improves performance for all tasks. For SR, using \abv is generally helpful with the exception of the Corn task, in which the SR base policy performed exceptionally well. For Eggplant, we found the SR-\abv policy to achieve only one fewer full completion than SR, but it achieved 6 more partial successes, showing its robust real-world performance. Overall, our results suggest that \abv can be a simple but effective way to improve retrieval across different, smooth latent representations. \looseness=-1

\subsection{Ablations}

\textbf{Importance Weights.} While \abv computes importance weights $\pt / \pprior$ as described in \cref{eq:our_method}, prior works only consider $\pt$. In \cref{tab:no_denom}, we ablate whether this normalization is important for \abv. Across simulated tasks, we generally find the normalization by $\pprior$ to be helpful, suggesting that using importance weights for retrieval leads to effective policy performance.

\begin{table}[]
\caption{We ablate removing the denominator $\pprior$ in \abv and halving the Gaussian KDE bandwidth parameter.}
\label{tab:no_denom}
\begin{adjustbox}{width=1\textwidth,center=\textwidth} 
\begin{tabular}{l|c|ccccc}
\toprule
\textbf{Method} & \textbf{Square} & \textbf{Mug-Micro.} & \textbf{Mug-Mug} & \textbf{Mug-Pudding} & \textbf{Soup-Cheese} & \textbf{Soup-Sauce} \\
\midrule
\abv & 84 \Std{2.8}  & 81 \Std{3.6} & 87 \Std{2.0} & 45 \Std{1.4} & 83 \Std{3.3} & 54 \Std{5.7}  \\
\abv (w/o norm) & 61 \Std{1.0} & 79 \Std{2.1} & 93 \Std{0.5} & 41 \Std{5.4} & 83 \Std{3.8} & 52 \Std{4.7} \\
\abv ($1/2$ bw) & -- & 79 \Std{1.4} & 92 \Std{1.6} & 40 \Std{0.9} & 79 \Std{1.4} & 56 \Std{1.6} \\
\bottomrule
\end{tabular}
\end{adjustbox}
\vspace{-0.2in}
\end{table}

\begin{wraptable}[8]{R}{0.2\textwidth}
\vspace{-2.5em}
\caption{\abv across retrieval thresholds.}
\label{tab:ret_percent}
\centering
\begin{tabular}{@{}cc@{}}
\toprule
\multicolumn{2}{c}{\textbf{Square}} \\
\% Ret    & Success        \\ \midrule
20        & 71 \Std{4.1}   \\
30        & 84 \Std{2.8}   \\
50        & 88 \Std{1.9}   \\
60        & 54 \Std{5.9}   \\ \bottomrule
\end{tabular}
\end{wraptable}

\textbf{Bandwidth Parameters.} We set the bandwidth factor for all Gaussian KDEs to multiplicative factor of Scott's rule, e.g. $h = c \times |\mathcal{D}|^{-1 / (d+4)}$ where $d$ is the dimension of the embedding vectors and $c$ is the multiplicative constant. The bottom row of \cref{tab:no_denom} shows the performance of \abv in LIBERO when we halve the multiplicative factor $c$ from 4 to 2. While performance is relatively robust, we find that smoother KDEs are slightly better on average.

\textbf{Retrieval Thresholds.} The amount of data retrieved can affect the policy performance. In \cref{tab:ret_percent}, we show that \abv's performance across different thresholds. Consistent with prior work, choosing the correct threshold is important. We see this effect in Square where a threshold of more than 50\% forces the retrieval of prior data from the incorrect task execution.  \looseness=-1

\section{Conclusion}
We introduce \fullname, an importance sampling-inspired method for retrieval. We find that \abv is able to better select data for retrieval, as evidenced by improved performance across both simulated and real tasks, including a long horizon task. Moreover, as \abv can easily be used with several retrieval works, we hope our insights will become standard practice for few-shot imitation learning. \looseness=-1

\textbf{Limitations.} While \abv can be effectively used on top of VAE-based latent spaces, it performs poorly on BYOL, a latent space without L2 smoothness constraints. We also do not further evaluate what makes an effective latent space, leaving this direction for future work. Moreover, due to the constraints of existing large scale prior datasets, our evaluation is largely limited to pick-place like tasks. Future work may explore retrieval for more complex and dexterous tasks. Finally, \abv assumes the use of Gaussian KDEs, which can become computational intractable and numerically unstable in higher dimensions, ultimately restricting the size of the latent representation that can be used for retrieval. Future work could seek to use more advanced methods for estimating importance weights.

\acknowledgments{We would like to thank members of ILIAD for their feedback and support. This work was funded by NSF grant \#2132847 and \#2218760, DARPA YFA, ONR YIP, and Cooperative AI foundation.}


\bibliography{references}  

\begin{thebibliography}{32}
\providecommand{\natexlab}[1]{#1}
\providecommand{\url}[1]{\texttt{#1}}
\expandafter\ifx\csname urlstyle\endcsname\relax
  \providecommand{\doi}[1]{doi: #1}\else
  \providecommand{\doi}{doi: \begingroup \urlstyle{rm}\Url}\fi

\bibitem[Radford et~al.(2021)Radford, Kim, Hallacy, Ramesh, Goh, Agarwal, Sastry, Askell, Mishkin, Clark, et~al.]{radford2021learning}
A.~Radford, J.~W. Kim, C.~Hallacy, A.~Ramesh, G.~Goh, S.~Agarwal, G.~Sastry, A.~Askell, P.~Mishkin, J.~Clark, et~al.
\newblock Learning transferable visual models from natural language supervision.
\newblock In \emph{International conference on machine learning}, pages 8748--8763. PMLR, 2021.

\bibitem[Reed et~al.(2022)Reed, Zolna, Parisotto, Colmenarejo, Novikov, Barth-maron, Gim{\'e}nez, Sulsky, Kay, Springenberg, Eccles, Bruce, Razavi, Edwards, Heess, Chen, Hadsell, Vinyals, Bordbar, and de~Freitas]{reed2022a}
S.~Reed, K.~Zolna, E.~Parisotto, S.~G. Colmenarejo, A.~Novikov, G.~Barth-maron, M.~Gim{\'e}nez, Y.~Sulsky, J.~Kay, J.~T. Springenberg, T.~Eccles, J.~Bruce, A.~Razavi, A.~Edwards, N.~Heess, Y.~Chen, R.~Hadsell, O.~Vinyals, M.~Bordbar, and N.~de~Freitas.
\newblock A generalist agent.
\newblock \emph{Transactions on Machine Learning Research}, 2022.
\newblock ISSN 2835-8856.
\newblock URL \url{https://openreview.net/forum?id=1ikK0kHjvj}.
\newblock Featured Certification, Outstanding Certification.

\bibitem[Zhao et~al.(2024)Zhao, Tompson, Driess, Florence, Ghasemipour, Finn, and Wahid]{zhao2024aloha}
T.~Z. Zhao, J.~Tompson, D.~Driess, P.~Florence, S.~K.~S. Ghasemipour, C.~Finn, and A.~Wahid.
\newblock {ALOHA} unleashed: A simple recipe for robot dexterity.
\newblock In \emph{8th Annual Conference on Robot Learning}, 2024.
\newblock URL \url{https://openreview.net/forum?id=gvdXE7ikHI}.

\bibitem[O’Neill et~al.(2024)O’Neill, Rehman, Maddukuri, Gupta, Padalkar, Lee, Pooley, Gupta, Mandlekar, Jain, et~al.]{openx}
A.~O’Neill, A.~Rehman, A.~Maddukuri, A.~Gupta, A.~Padalkar, A.~Lee, A.~Pooley, A.~Gupta, A.~Mandlekar, A.~Jain, et~al.
\newblock Open x-embodiment: Robotic learning datasets and rt-x models: Open x-embodiment collaboration 0.
\newblock In \emph{2024 IEEE International Conference on Robotics and Automation (ICRA)}, pages 6892--6903. IEEE, 2024.

\bibitem[Du et~al.(2023)Du, Nair, Sadigh, and Finn]{du2023behavior}
M.~Du, S.~Nair, D.~Sadigh, and C.~Finn.
\newblock Behavior retrieval: Few-shot imitation learning by querying unlabeled datasets.
\newblock \emph{arXiv preprint arXiv:2304.08742}, 2023.

\bibitem[Lin et~al.()Lin, Cui, Xie, Hua, and Sadigh]{lin2024flowretrieval}
L.-H. Lin, Y.~Cui, A.~Xie, T.~Hua, and D.~Sadigh.
\newblock Flowretrieval: Flow-guided data retrieval for few-shot imitation learning.
\newblock In \emph{8th Annual Conference on Robot Learning}.

\bibitem[Nasiriany et~al.(2022)Nasiriany, Gao, Mandlekar, and Zhu]{nasiriany2022sailor}
S.~Nasiriany, T.~Gao, A.~Mandlekar, and Y.~Zhu.
\newblock Learning and retrieval from prior data for skill-based imitation learning.
\newblock In \emph{Conference on Robot Learning (CoRL)}, 2022.

\bibitem[Walke et~al.(2023)Walke, Black, Lee, Kim, Du, Zheng, Zhao, Hansen-Estruch, Vuong, He, Myers, Fang, Finn, and Levine]{bridge}
H.~Walke, K.~Black, A.~Lee, M.~J. Kim, M.~Du, C.~Zheng, T.~Zhao, P.~Hansen-Estruch, Q.~Vuong, A.~He, V.~Myers, K.~Fang, C.~Finn, and S.~Levine.
\newblock Bridgedata v2: A dataset for robot learning at scale.
\newblock In \emph{Conference on Robot Learning (CoRL)}, 2023.

\bibitem[Khazatsky et~al.(2024)Khazatsky, Pertsch, Nair, Balakrishna, Dasari, Karamcheti, Nasiriany, Srirama, Chen, Ellis, et~al.]{khazatsky2024droid}
A.~Khazatsky, K.~Pertsch, S.~Nair, A.~Balakrishna, S.~Dasari, S.~Karamcheti, S.~Nasiriany, M.~K. Srirama, L.~Y. Chen, K.~Ellis, et~al.
\newblock Droid: A large-scale in-the-wild robot manipulation dataset.
\newblock \emph{arXiv preprint arXiv:2403.12945}, 2024.

\bibitem[Memmel et~al.(2025)Memmel, Berg, Chen, Gupta, and Francis]{memmel2025strap}
M.~Memmel, J.~Berg, B.~Chen, A.~Gupta, and J.~Francis.
\newblock {STRAP}: Robot sub-trajectory retrieval for augmented policy learning.
\newblock In \emph{The Thirteenth International Conference on Learning Representations}, 2025.
\newblock URL \url{https://openreview.net/forum?id=4VHiptx7xe}.

\bibitem[Yue et~al.(2024)Yue, Liu, Hua, Ren, Lin, Zhang, and Zhang]{yue2024leverage}
S.~Yue, J.~Liu, X.~Hua, J.~Ren, S.~Lin, J.~Zhang, and Y.~Zhang.
\newblock How to leverage diverse demonstrations in offline imitation learning.
\newblock \emph{arXiv preprint arXiv:2405.17476}, 2024.

\bibitem[Hanna et~al.(2019)Hanna, Niekum, and Stone]{hanna2019importance}
J.~Hanna, S.~Niekum, and P.~Stone.
\newblock Importance sampling policy evaluation with an estimated behavior policy.
\newblock In \emph{International Conference on Machine Learning}, pages 2605--2613. PMLR, 2019.

\bibitem[Sinha et~al.(2022)Sinha, Song, Garg, and Ermon]{sinha2022experience}
S.~Sinha, J.~Song, A.~Garg, and S.~Ermon.
\newblock Experience replay with likelihood-free importance weights.
\newblock In \emph{Learning for Dynamics and Control Conference}, pages 110--123. PMLR, 2022.

\bibitem[Xie et~al.(2023)Xie, Santurkar, Ma, and Liang]{xie2023data}
S.~M. Xie, S.~Santurkar, T.~Ma, and P.~S. Liang.
\newblock Data selection for language models via importance resampling.
\newblock \emph{Advances in Neural Information Processing Systems}, 36:\penalty0 34201--34227, 2023.

\bibitem[Bennett(1976)]{BENNETT1976245}
C.~H. Bennett.
\newblock Efficient estimation of free energy differences from monte carlo data.
\newblock \emph{Journal of Computational Physics}, 22\penalty0 (2):\penalty0 245--268, 1976.
\newblock ISSN 0021-9991.
\newblock \doi{https://doi.org/10.1016/0021-9991(76)90078-4}.
\newblock URL \url{https://www.sciencedirect.com/science/article/pii/0021999176900784}.

\bibitem[Rhodes et~al.(2020)Rhodes, Xu, and Gutmann]{NEURIPS2020_33d3b157}
B.~Rhodes, K.~Xu, and M.~U. Gutmann.
\newblock Telescoping density-ratio estimation.
\newblock In H.~Larochelle, M.~Ranzato, R.~Hadsell, M.~Balcan, and H.~Lin, editors, \emph{Advances in Neural Information Processing Systems}, volume~33, pages 4905--4916. Curran Associates, Inc., 2020.
\newblock URL \url{https://proceedings.neurips.cc/paper_files/paper/2020/file/33d3b157ddc0896addfb22fa2a519097-Paper.pdf}.

\bibitem[Choi et~al.(2022)Choi, Meng, Song, and Ermon]{choi2022density}
K.~Choi, C.~Meng, Y.~Song, and S.~Ermon.
\newblock Density ratio estimation via infinitesimal classification.
\newblock In \emph{International Conference on Artificial Intelligence and Statistics}, pages 2552--2573. PMLR, 2022.

\bibitem[Maddukuri et~al.(2025)Maddukuri, Jiang, Chen, Nasiriany, Xie, Fang, Huang, Wang, Xu, Chernyadev, Reed, Goldberg, Mandlekar, Fan, and Zhu]{maddukuri2025simandrealcotrainingsimplerecipe}
A.~Maddukuri, Z.~Jiang, L.~Y. Chen, S.~Nasiriany, Y.~Xie, Y.~Fang, W.~Huang, Z.~Wang, Z.~Xu, N.~Chernyadev, S.~Reed, K.~Goldberg, A.~Mandlekar, L.~Fan, and Y.~Zhu.
\newblock Sim-and-real co-training: A simple recipe for vision-based robotic manipulation, 2025.
\newblock URL \url{https://arxiv.org/abs/2503.24361}.

\bibitem[Hejna et~al.(2024)Hejna, Bhateja, Jiang, Pertsch, and Sadigh]{hejna2024remix}
J.~Hejna, C.~A. Bhateja, Y.~Jiang, K.~Pertsch, and D.~Sadigh.
\newblock Remix: Optimizing data mixtures for large scale imitation learning.
\newblock In \emph{8th Annual Conference on Robot Learning}, 2024.
\newblock URL \url{https://openreview.net/forum?id=fIj88Tn3fc}.

\bibitem[Belkhale et~al.(2024)Belkhale, Cui, and Sadigh]{belkhale2024data}
S.~Belkhale, Y.~Cui, and D.~Sadigh.
\newblock Data quality in imitation learning.
\newblock \emph{Advances in Neural Information Processing Systems}, 36, 2024.

\bibitem[Hejna et~al.(2025)Hejna, Mirchandani, Balakrishna, Xie, Wahid, Tompson, Sanketi, Shah, Devin, and Sadigh]{hejna2025robot}
J.~Hejna, S.~Mirchandani, A.~Balakrishna, A.~Xie, A.~Wahid, J.~Tompson, P.~Sanketi, D.~Shah, C.~Devin, and D.~Sadigh.
\newblock Robot data curation with mutual information estimators.
\newblock \emph{arXiv preprint arXiv:2502.08623}, 2025.

\bibitem[Kuhar et~al.(2023)Kuhar, Cheng, Chopra, Bronars, and Xu]{kuhar2023learning}
S.~Kuhar, S.~Cheng, S.~Chopra, M.~Bronars, and D.~Xu.
\newblock Learning to discern: Imitating heterogeneous human demonstrations with preference and representation learning.
\newblock In \emph{7th Annual Conference on Robot Learning}, 2023.

\bibitem[Beliaev et~al.(2022)Beliaev, Shih, Ermon, Sadigh, and Pedarsani]{beliaev2022imitation}
M.~Beliaev, A.~Shih, S.~Ermon, D.~Sadigh, and R.~Pedarsani.
\newblock Imitation learning by estimating expertise of demonstrators.
\newblock In \emph{International Conference on Machine Learning}, pages 1732--1748. PMLR, 2022.

\bibitem[Ha et~al.(2023)Ha, Florence, and Song]{ha2023scalingup}
H.~Ha, P.~Florence, and S.~Song.
\newblock Scaling up and distilling down: Language-guided robot skill acquisition.
\newblock In \emph{Proceedings of the 2023 Conference on Robot Learning}, 2023.

\bibitem[Mandlekar et~al.(2023)Mandlekar, Nasiriany, Wen, Akinola, Narang, Fan, Zhu, and Fox]{mandlekar2023mimicgen}
A.~Mandlekar, S.~Nasiriany, B.~Wen, I.~Akinola, Y.~Narang, L.~Fan, Y.~Zhu, and D.~Fox.
\newblock Mimicgen: A data generation system for scalable robot learning using human demonstrations.
\newblock In \emph{7th Annual Conference on Robot Learning}, 2023.

\bibitem[Scott(2015)]{scott2015multivariate}
D.~W. Scott.
\newblock \emph{Multivariate density estimation: theory, practice, and visualization}.
\newblock John Wiley \& Sons, 2015.

\bibitem[Gelman and Meng(2004)]{gelman2004applied}
A.~Gelman and X.-L. Meng.
\newblock \emph{Applied Bayesian modeling and causal inference from incomplete-data perspectives}.
\newblock John Wiley \& Sons, 2004.

\bibitem[Mandlekar et~al.(2021)Mandlekar, Xu, Wong, Nasiriany, Wang, Kulkarni, Fei-Fei, Savarese, Zhu, and Mart\'{i}n-Mart\'{i}n]{robomimic2021}
A.~Mandlekar, D.~Xu, J.~Wong, S.~Nasiriany, C.~Wang, R.~Kulkarni, L.~Fei-Fei, S.~Savarese, Y.~Zhu, and R.~Mart\'{i}n-Mart\'{i}n.
\newblock What matters in learning from offline human demonstrations for robot manipulation.
\newblock In \emph{Conference on Robot Learning (CoRL)}, 2021.

\bibitem[Liu et~al.(2023)Liu, Zhu, Gao, Feng, Liu, Zhu, and Stone]{liu2023libero}
B.~Liu, Y.~Zhu, C.~Gao, Y.~Feng, Q.~Liu, Y.~Zhu, and P.~Stone.
\newblock Libero: Benchmarking knowledge transfer for lifelong robot learning.
\newblock \emph{arXiv preprint arXiv:2306.03310}, 2023.

\bibitem[Chi et~al.(2023)Chi, Xu, Feng, Cousineau, Du, Burchfiel, Tedrake, and Song]{chi2023diffusion}
C.~Chi, Z.~Xu, S.~Feng, E.~Cousineau, Y.~Du, B.~Burchfiel, R.~Tedrake, and S.~Song.
\newblock Diffusion policy: Visuomotor policy learning via action diffusion.
\newblock \emph{The International Journal of Robotics Research}, page 02783649241273668, 2023.

\bibitem[Pari et~al.(2021)Pari, Shafiullah, Arunachalam, and Pinto]{pari2021surprising}
J.~Pari, N.~M. Shafiullah, S.~P. Arunachalam, and L.~Pinto.
\newblock The surprising effectiveness of representation learning for visual imitation, 2021.

\bibitem[Grill et~al.(2020)Grill, Strub, Altché, Tallec, Richemond, Buchatskaya, Doersch, Pires, Guo, Azar, Piot, Kavukcuoglu, Munos, and Valko]{grill2020bootstraplatentnewapproach}
J.-B. Grill, F.~Strub, F.~Altché, C.~Tallec, P.~H. Richemond, E.~Buchatskaya, C.~Doersch, B.~A. Pires, Z.~D. Guo, M.~G. Azar, B.~Piot, K.~Kavukcuoglu, R.~Munos, and M.~Valko.
\newblock Bootstrap your own latent: A new approach to self-supervised learning, 2020.
\newblock URL \url{https://arxiv.org/abs/2006.07733}.

\end{thebibliography}
\clearpage
\appendix
\section{Website}
Please find code and videos of sample task rollouts and more on our site: \url{https://rahulschand.github.io/iwr/}.

\section{Implementation Details}

\textbf{Hyperparameters.} We provide hyper-parameters for all retrieval methods in \cref{tab:vae_hyperparameters} and for Diffusion Policy \citep{chi2023diffusion}, which was used for all policy learning evaluations, in \cref{tab:dp_hyperparameters}. While we found the original hyper-parameters from \citet{lin2024flowretrieval, du2023behavior} to work well for Robomimic and Bridge, we found that they did not perform well for LIBERO, likely due to the use of large action chunks. We thus modified the VAE to accept action chunks, use state history, and down-weight image reconstruction which lead to overall better performance for all methods and baselines. This obviated the need to additionally append the action to the learned representation $z$ as done by  \citet{lin2024flowretrieval, du2023behavior}. 

\textbf{Architectures.} For all VAEs we use a ResNet18 encoder-decoder architecture, with MLPs in between to process the concatenated image embeddings, robot proprioceptive state, and actions. For LIBERO only, we additionally use a small 2 layer transformer for the action encoder and decoder instead of projection layers from the MLP.

\textbf{Simulation Evaluation Procedure.} For evaluating policies in sim we run three seeds for 100k timesteps. We evaluate each policy every 25K steps for 50 episodes. Following the evaluation procedure of \citet{robomimic2021, chi2023diffusion}, we take the average of the best performing checkpoint across all seeds. 

\textbf{Real Evaluation Procedure.} For real world evaluations we train a single policy for a fixed number of timesteps and run 20 evaluation trials.

\begin{table}[H]
\centering
\caption{Hyperparameters used for retrieval methods.}
\label{tab:vae_hyperparameters}
\begin{tabular}{@{}ccccc@{}}
\toprule
Method                  & Parameter          & Robomimic & LIBERO     & Bridge     \\ \midrule
\multirow{8}{*}{Shared} & Optimizer          & \multicolumn{3}{c}{Adam}            \\
                        & Learning Rate      & \multicolumn{3}{c}{0.0001}          \\
                        & Batch Size         & \multicolumn{3}{c}{256}             \\
                        & Training Steps     & 200,000   & 200,000    & 400,000    \\
                        & Image Resolution   & (84, 84)  & (128, 128) & (224, 224) \\
                        & Augmentations      & \multicolumn{3}{c}{None}            \\
                        & $\alpha$            & \multicolumn{3}{c}{0.5}          \\
                        & $\beta$            & \multicolumn{3}{c}{0.0001}          \\
                        & $z$ dim            & 16        & 32         & 32         \\ \midrule
\multirow{4}{*}{BR}     & Image Recon Weight & 1         & 0.01       & 1          \\
                        & Action Chunk       & 1         & 16         & 4          \\
                        & Append Action      & TRUE      & FALSE      & TRUE       \\
                        & State History      & 1         & 2          & 1          \\ \midrule
\multirow{5}{*}{FR}     & Steps between Flow Frames  & 8 & 8 & 8        \\
                        & Image Recon Weight & 1         & 0.01       & 1          \\
                        & Action Chunk       & 1         & 16         & 4          \\
                        & Append Action      & TRUE      & FALSE      & TRUE       \\
                        & State History      & 1         & 2          & 1          \\ \midrule
\multirow{4}{*}{SAILOR} & Obs \& Action Chunk & \multicolumn{3}{c}{10}              \\
                        & Time Loss Weight   & \multicolumn{3}{c}{0.000001}        \\
                        & Training Steps   & \multicolumn{3}{c}{200,000}        \\
                        & Max Seq Offset     & \multicolumn{3}{c}{50}              \\ \bottomrule
\end{tabular}
\end{table}

\begin{table}[H]
\caption{Diffusion Policy hyperparameters used for policy learning evaluations.}
\label{tab:dp_hyperparameters}
\centering
\begin{tabular}{@{}cc@{}}
\toprule
Parameter        & Value                             \\ \midrule
Optimizer        & Adam                              \\
Learning Rate    & 0.0001                            \\
Batch Size       & 256                               \\
Training Steps   & 100,000                           \\
Obs History      & 2                                 \\
Action Chunk     & 16                                \\
Image Resolution & See \cref{tab:vae_hyperparameters}                   \\
Augmentations    & Random Scale and Crop (0.85, 1.0) \\ \bottomrule
\end{tabular}
\end{table}

\section{Additional Experiments}
We include additional ablations in \cref{tab:tasks}, 
\cref{tab:strap}, \cref{tab:bandwidth}, and \cref{tab:byol}. To train our BYOL model, we use the codebase from VINN~\citep{pari2021surprising} and train for 100000 steps. We remove grayscale and random horizontal flip data augmentations due to training instabilities.

\begin{table}[h]
\caption{We compare \abv with BR for the remaining 5 LIBERO tasks.}
\label{tab:tasks}
\centering
\begin{tabular}{l|ccccc}
\toprule
\textbf{Method} & \textbf{Book-} & \textbf{Bowl-} & \textbf{Cream-} & \textbf{Pots-} & \textbf{Stove-}  \\
& \textbf{Caddy} & \textbf{Cabinet} & \textbf{Cheese} & \textbf{Stove} & \textbf{Moka} \\
\midrule
BR & 97.3 \Std{1.4} & 99.3 \Std{0.5} & 84.7 \Std{2.0} & 71.3 \Std{0.5} & 80.7 \Std{0.5} \\
\abv & 100.0 \Std{0.0} & 97.3 \Std{0.5} & 92.0 \Std{1.6} & 80.7 \Std{2.0} & 87.3 \Std{3.3}\\

\bottomrule
\end{tabular}
\end{table}

\begin{table}[h]
\caption{We compare against STRAP~\citep{memmel2025strap}, a trajectory-based, dynamic time-warping retrieval method.}
\label{tab:strap}
\centering
\begin{tabular}{l|c|ccccc|}
\toprule
\textbf{Method} & \textbf{Square} & \textbf{Mug-} & \textbf{Mug-} & \textbf{Mug-} & \textbf{Soup-} & \textbf{Soup-}  \\
& & \textbf{Microwave} & \textbf{Mug} & \textbf{Pudding} & \textbf{Cheese} & \textbf{Sauce} \\
\midrule
STRAP & 44.7 \Std{6.8} & \textbf{80.0} \Std{1.9} & 72.7 \Std{1.1} & 18.7 \Std{3.9} & 70.0 \Std{1.6} & 33.3 \Std{4.4} \\
\abv & \textbf{84} \Std{2.8}  & \textbf{81} \Std{3.6} & {87} \Std{2.0} & \textbf{45} \Std{1.4} & \textbf{83} \Std{3.3} & \textbf{54} \Std{5.7} \\

\bottomrule
\end{tabular}
\end{table}

\begin{table}[h]
\caption{\abv-0.5$h$ shows the effect of halving the bandwidth parameter $h$. We fix $h$ to be a multiplicative factor of Scott's rule, and we find \abv to be robust to $h$.}
\label{tab:bandwidth}
\centering
\begin{tabular}{l|c|ccccc|}
\toprule
\textbf{Method} & \textbf{Square} & \textbf{Mug-} & \textbf{Mug-} & \textbf{Mug-} & \textbf{Soup-} & \textbf{Soup-}  \\
& & \textbf{Microwave} & \textbf{Mug} & \textbf{Pudding} & \textbf{Cheese} & \textbf{Sauce} \\
\midrule
\abv-0.5$h$ & - & \textbf{79.3} \Std{1.4} & \textbf{92.0} \Std{1.6} & 40.0 \Std{0.9} & 79.3 \Std{1.4} & \textbf{56.0} \Std{1.6} \\
\abv & \textbf{84} \Std{2.8}  & \textbf{81} \Std{3.6} & {87} \Std{2.0} & \textbf{45} \Std{1.4} & \textbf{83} \Std{3.3} & \textbf{54} \Std{5.7} \\

\bottomrule
\end{tabular}
\end{table}

\begin{table}[h]
\caption{We compare against different learned latent spaces. BYOL~\citep{grill2020bootstraplatentnewapproach} is an alternate representation space shown to be useful for imitation models via nearest-neighbor retrieval in ~\citep{pari2021surprising}. BYOL does not have the same smoothness constraints as VAEs, leading to worse policy performance when retrieving according to BYOL latents. In addition, we find \abv to lead to minimal improvements over this latent space. BYOL-ResNet refers to the 2048-dimensional ResNet embedding, and BYOL-Proj refers to the 64-dimensional projection.}
\label{tab:byol}
\centering
\begin{tabular}{l|c|ccccc|}
\toprule
\textbf{Method} & \textbf{Square} & \textbf{Mug-} & \textbf{Mug-} & \textbf{Mug-} & \textbf{Soup-} & \textbf{Soup-}  \\
& & \textbf{Microwave} & \textbf{Mug} & \textbf{Pudding} & \textbf{Cheese} & \textbf{Sauce} \\
\midrule
BYOL-ResNet & - & 75.3 \Std{3.0} & 88.0 \Std{0.9} & 31.3 \Std{2.4}  & 57.3 \Std{7.1} & 48.0 \Std{3.8} \\
BYOL-Proj & - & 71.3 \Std{0.5} & 89.3 \Std{2.4} & 26.0 \Std{1.4}&  60.0 \Std{12.7} & 45.3 \Std{5.2} \\
BYOL-Proj-\abv & - & 74.0 \Std{0.9} & 84.0 \Std{2.5} & 28.7 \Std{6.1} &  48.0 \Std{8.6} & 44.0 \Std{2.5} \\
\midrule
\abv & \textbf{84} \Std{2.8}  & \textbf{81} \Std{3.6} & {87} \Std{2.0} & \textbf{45} \Std{1.4} & \textbf{83} \Std{3.3} & \textbf{54} \Std{5.7} \\

\bottomrule
\end{tabular}
\end{table}

\section{Additional Visualizations}

\begin{figure}[h]
    \centering
    \includegraphics[width=0.49\linewidth]{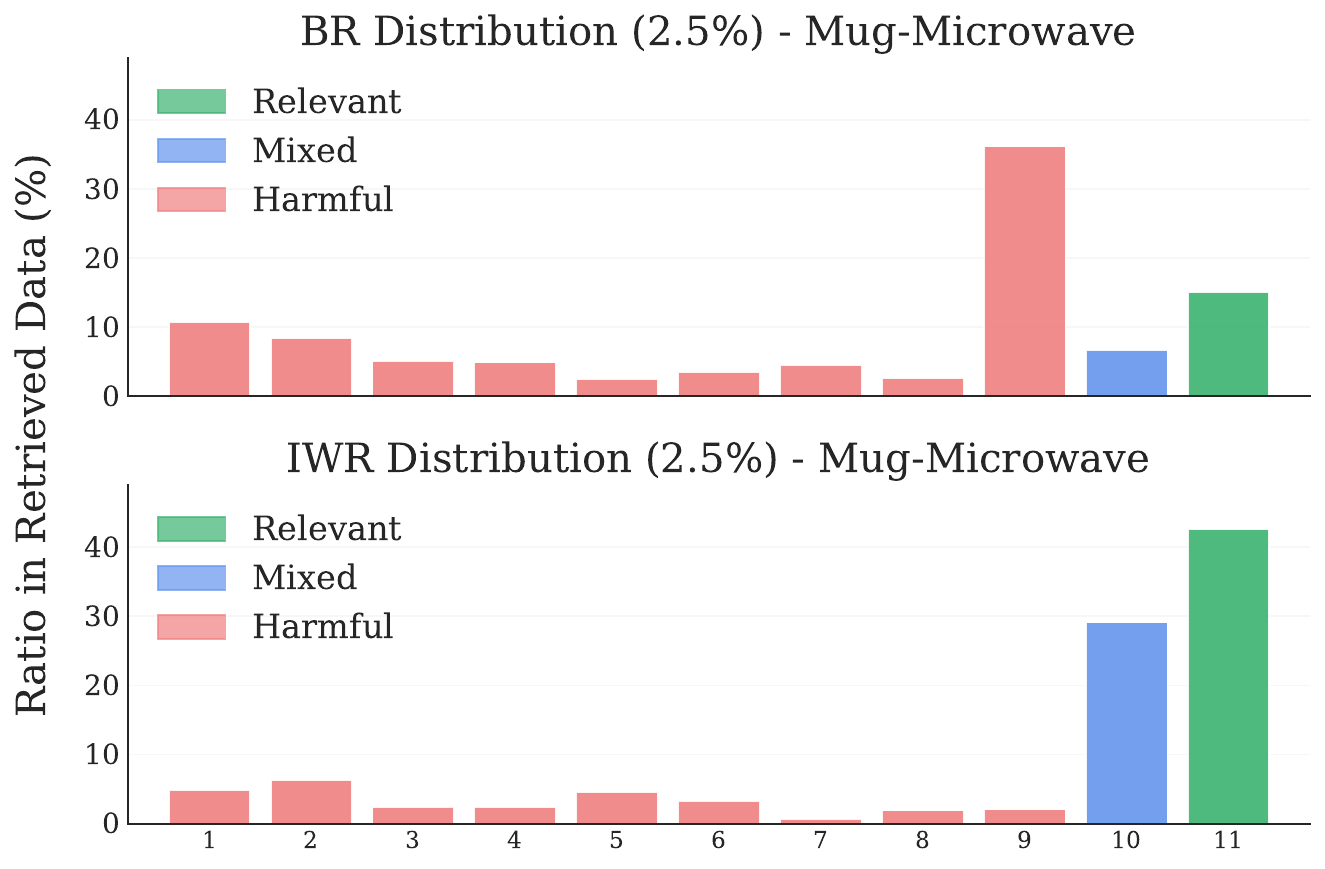}
    \includegraphics[width=0.49\linewidth]{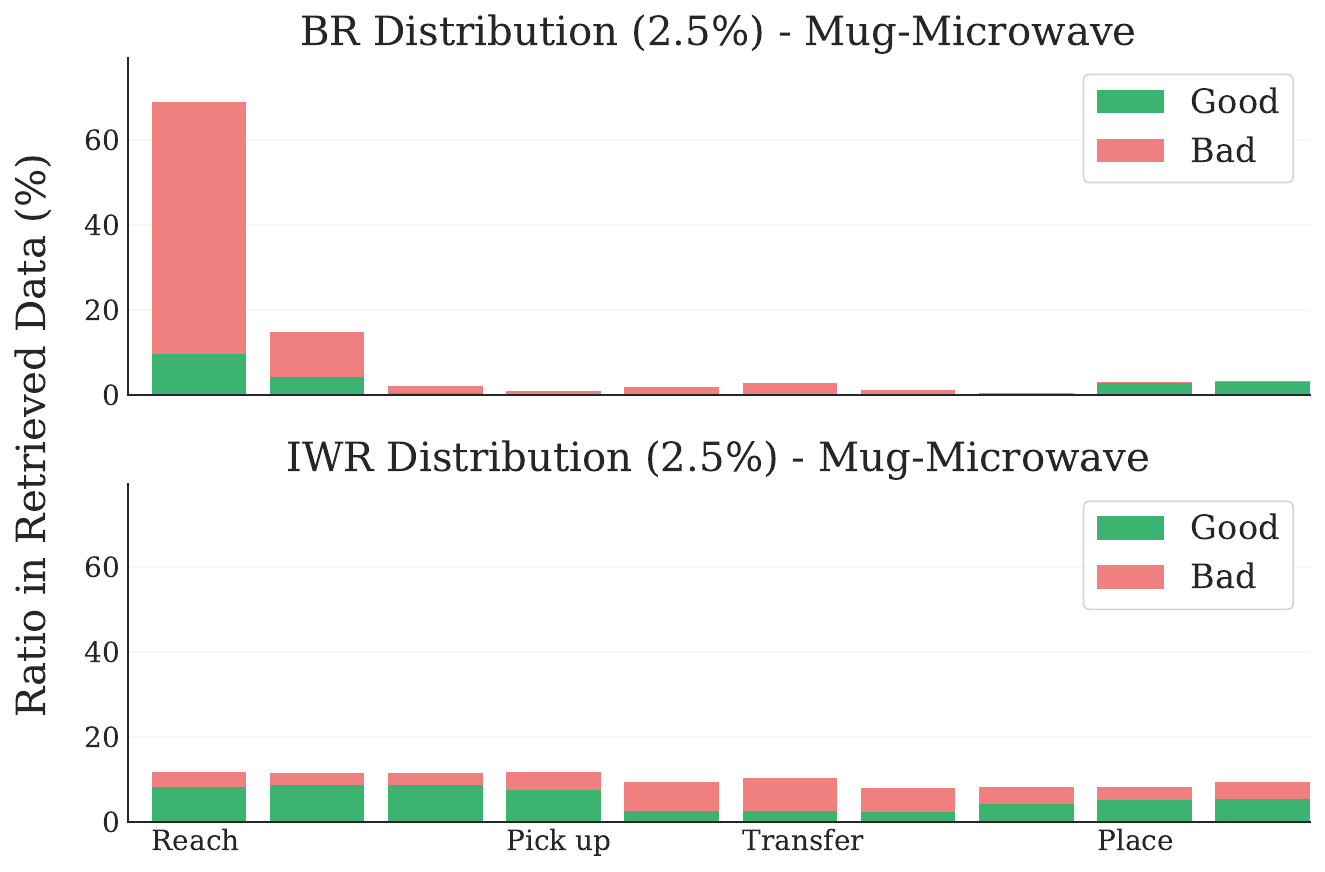}

    \includegraphics[width=\linewidth]{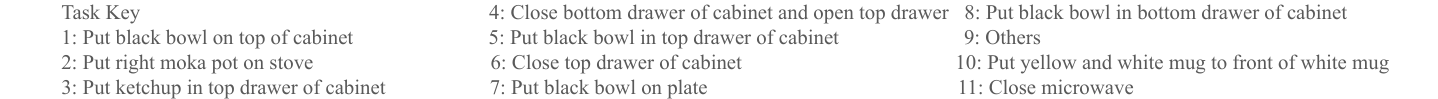}
    \caption{\textbf{Mug-Microwave LIBERO Task.} \textbf{(Left)} Retrieval distribution across tasks. \textbf{(Right)} Retrieval distribution across timesteps.}
    \label{fig:tasks-ret-1}
    \vspace{-0.25in}
\end{figure}

\begin{figure}[h]
    \centering
    \includegraphics[width=0.49\linewidth]{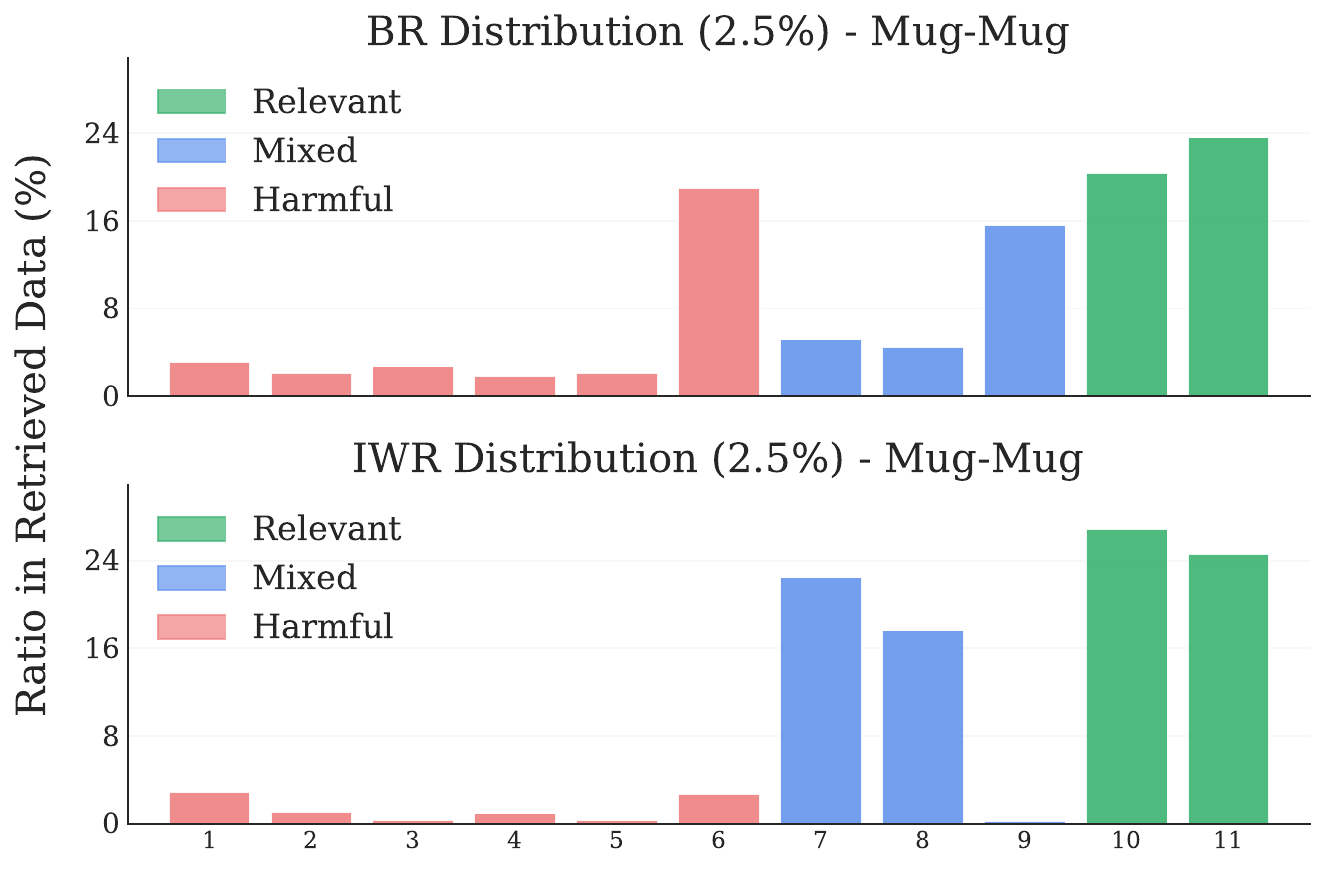}
    \includegraphics[width=0.49\linewidth]{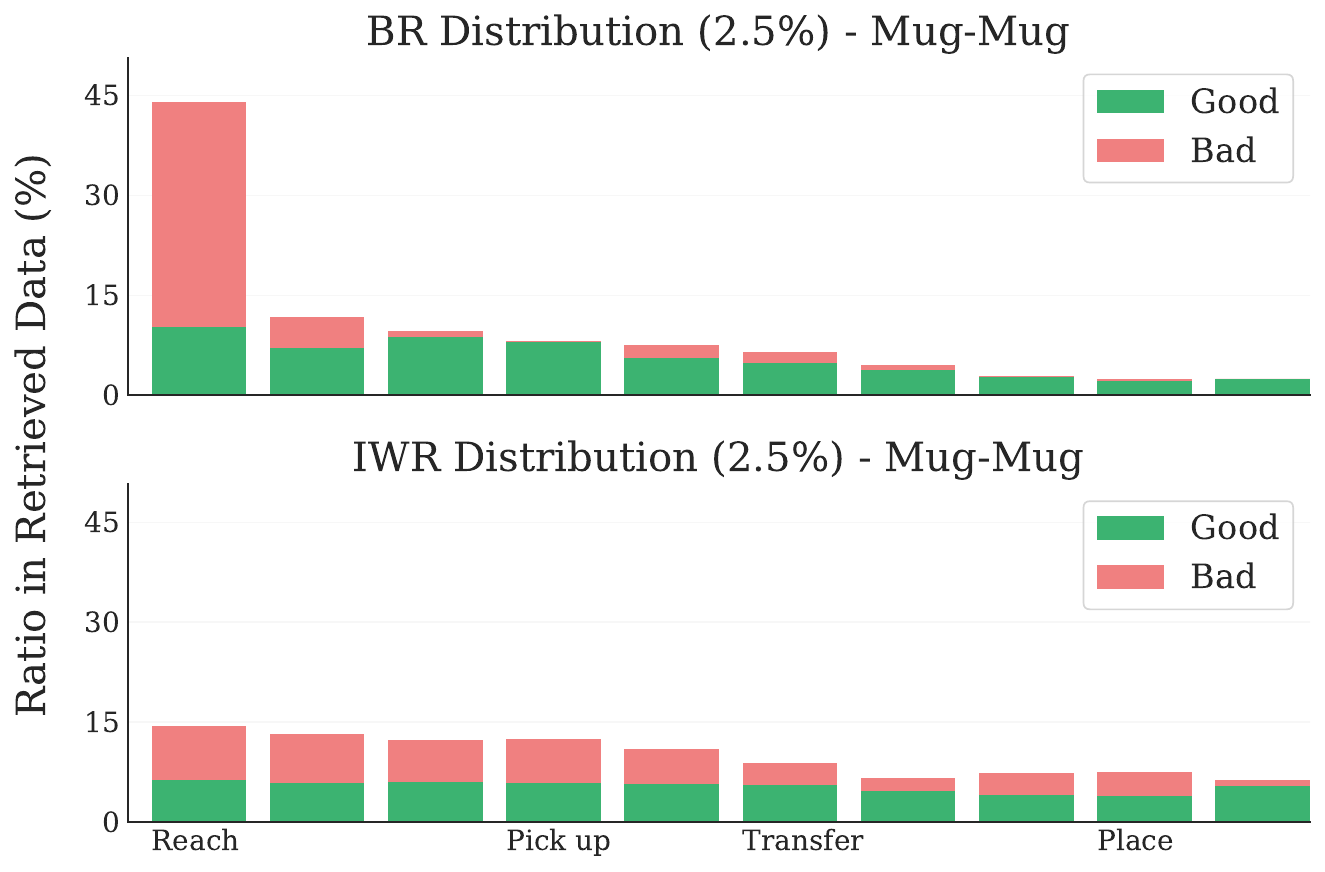}

    \includegraphics[width=\linewidth]{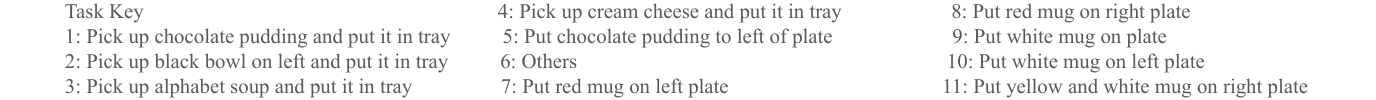}
    \caption{\textbf{Mug-Mug LIBERO Task.} \textbf{(Left)} Retrieval distribution across tasks. \textbf{(Right)} Retrieval distribution across timesteps.}
    \label{fig:tasks-ret-2}
\end{figure}

\begin{figure}[h]
    \centering
    \includegraphics[width=0.49\linewidth]{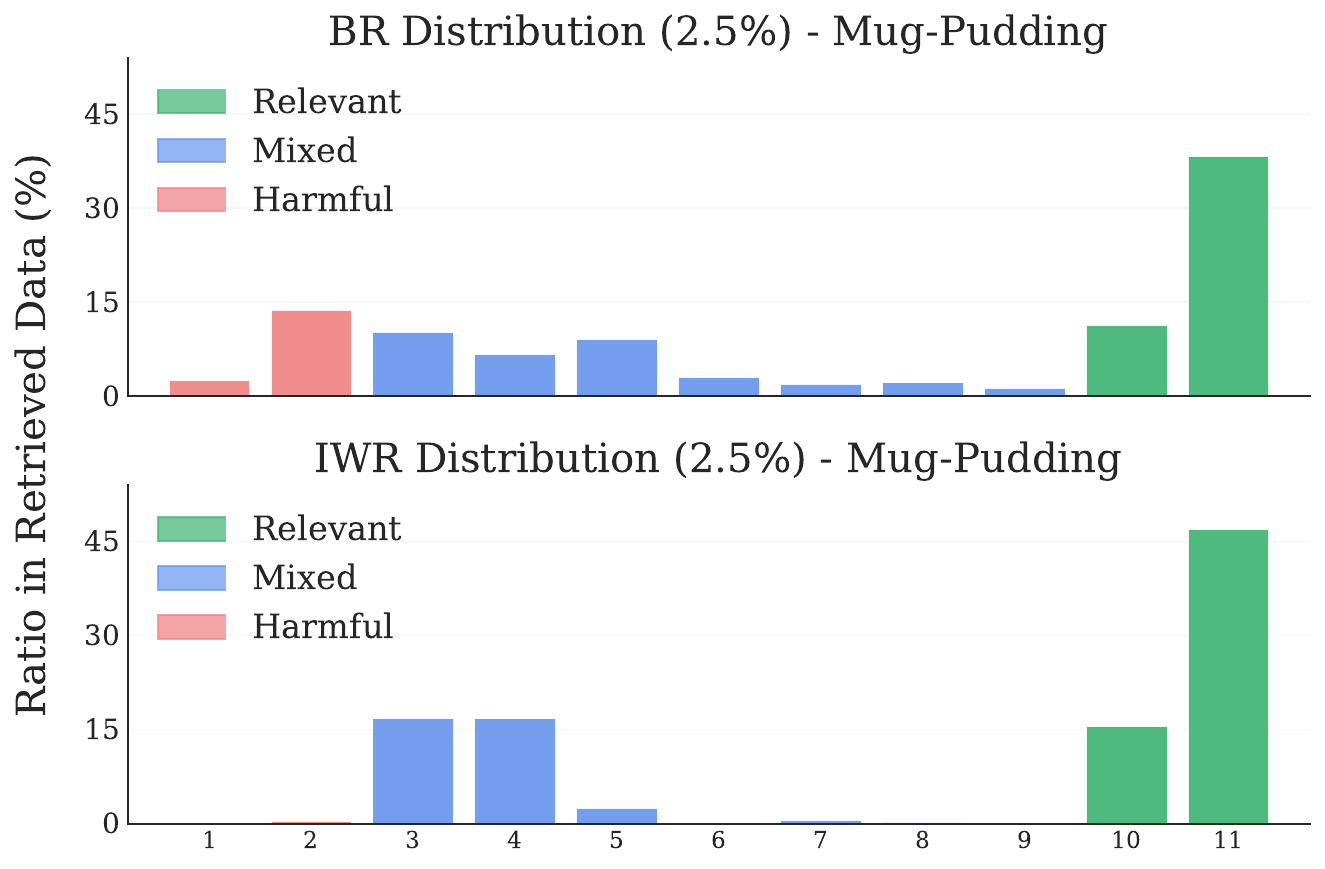}
    \includegraphics[width=0.49\linewidth]{retrieval_plots/steps/mug_pudding_step_stacked.pdf}

    \includegraphics[width=\linewidth]{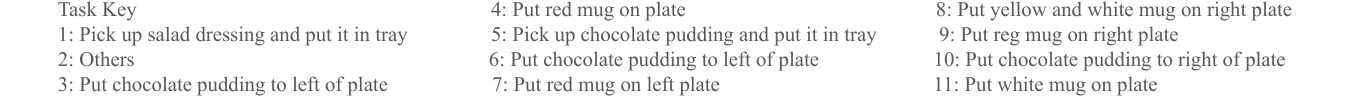}
    \caption{\textbf{Mug-Pudding LIBERO Task.} \textbf{(Left)} Retrieval distribution across tasks. \textbf{(Right)} Retrieval distribution across timesteps.}
    \label{fig:tasks-ret-3}
    \vspace{-0.25in}
\end{figure}

\begin{figure}[h]
    \centering
    \includegraphics[width=0.49\linewidth]{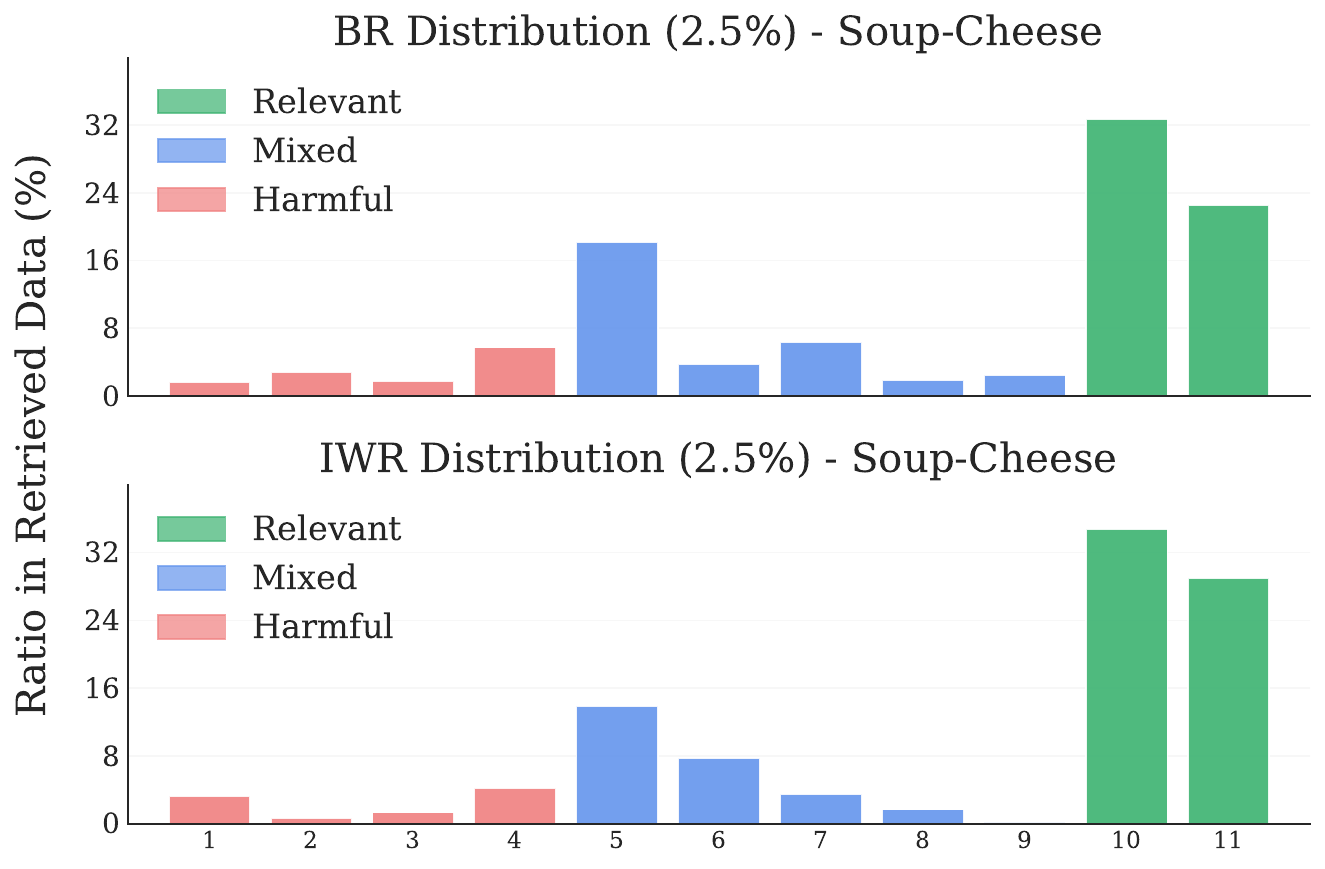}
    \includegraphics[width=0.49\linewidth]{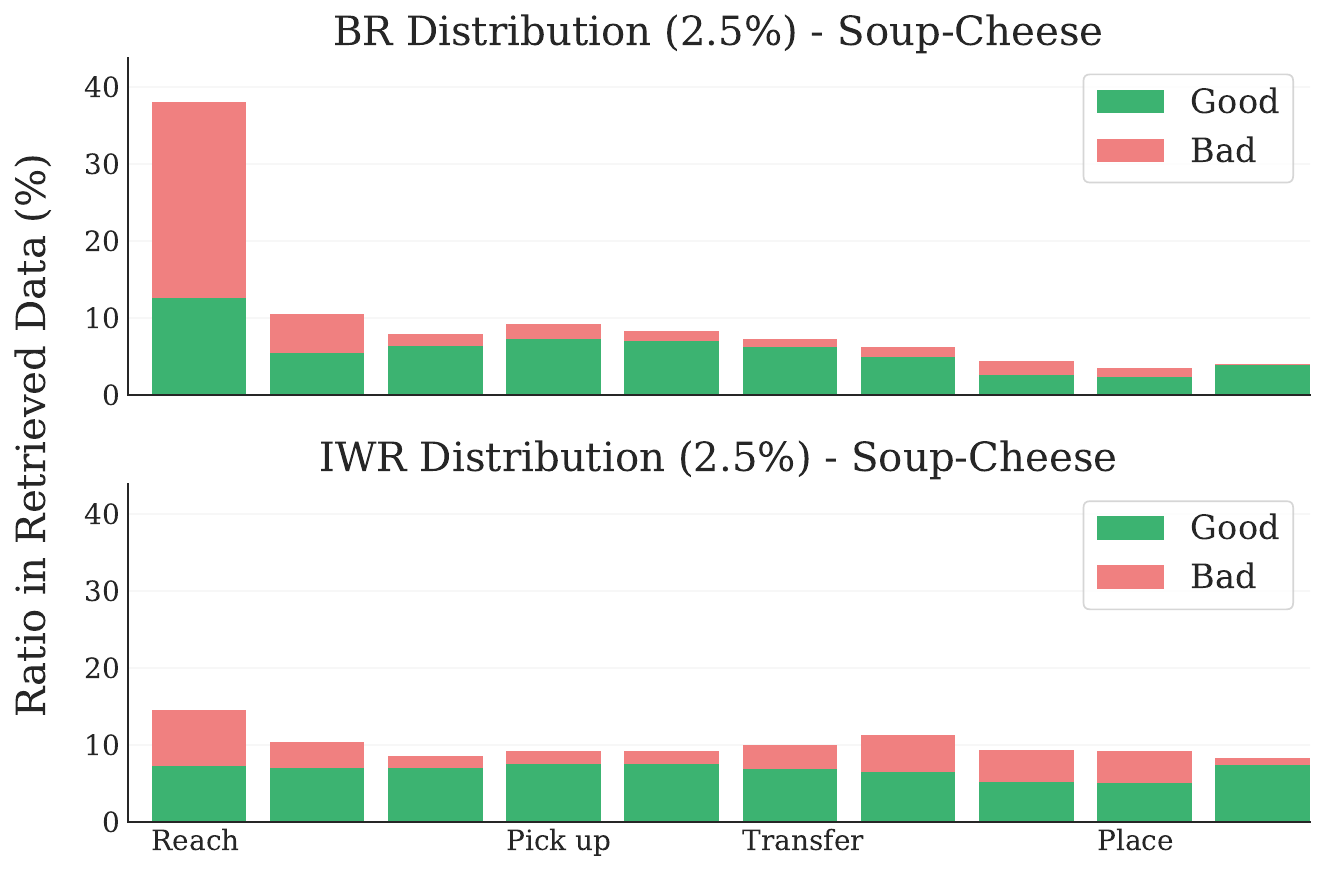}

    \includegraphics[width=\linewidth]{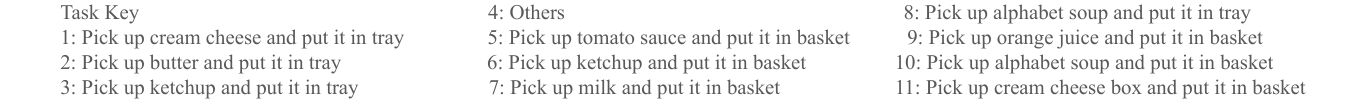}
    \caption{\textbf{Soup-Cheese LIBERO Task.} \textbf{(Left)} Retrieval distribution across tasks. \textbf{(Right)} Retrieval distribution across timesteps.}
    \label{fig:tasks-ret-4}
\end{figure}

\begin{figure}[h]
    \centering
    \includegraphics[width=0.49\linewidth]{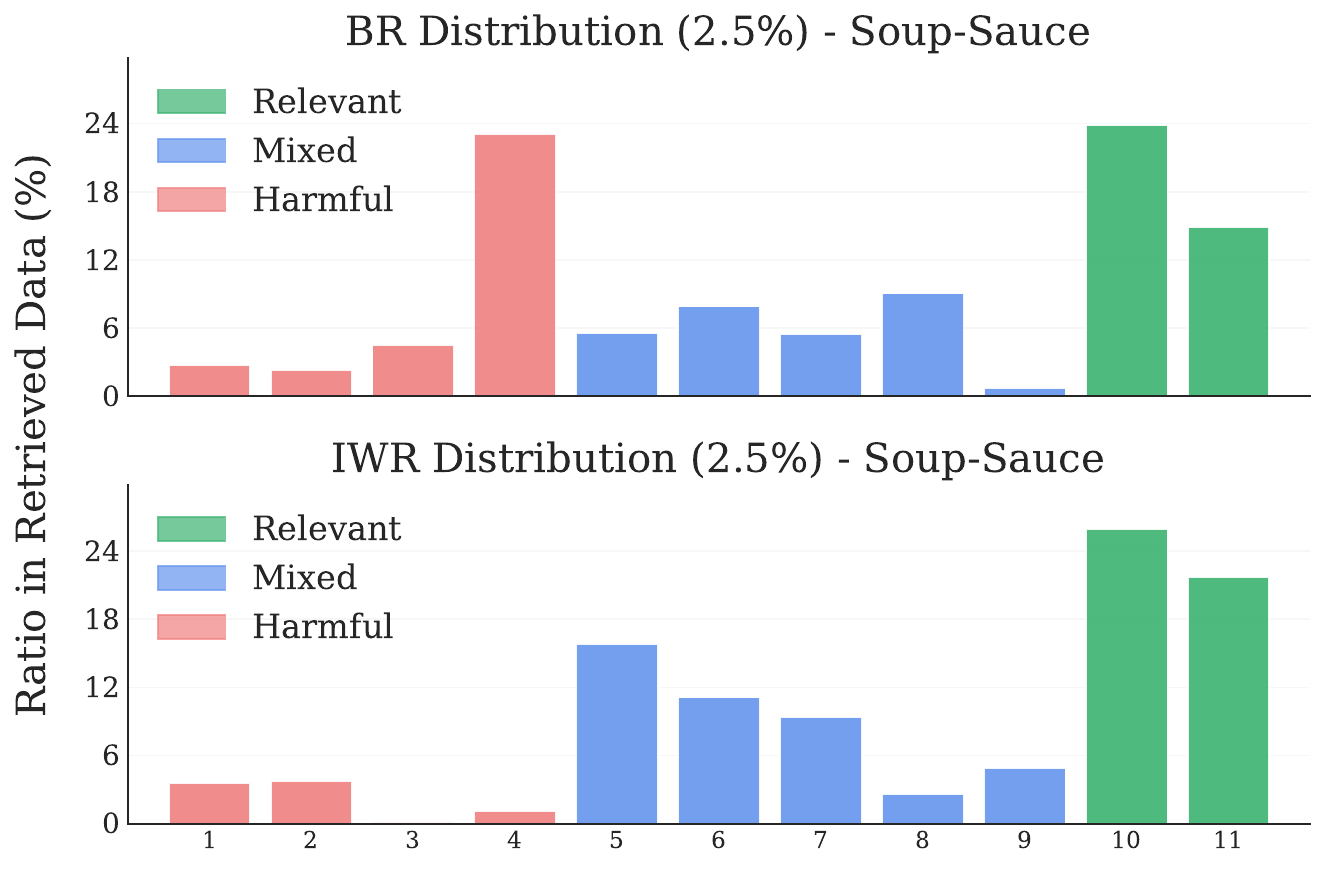}
    \includegraphics[width=0.49\linewidth]{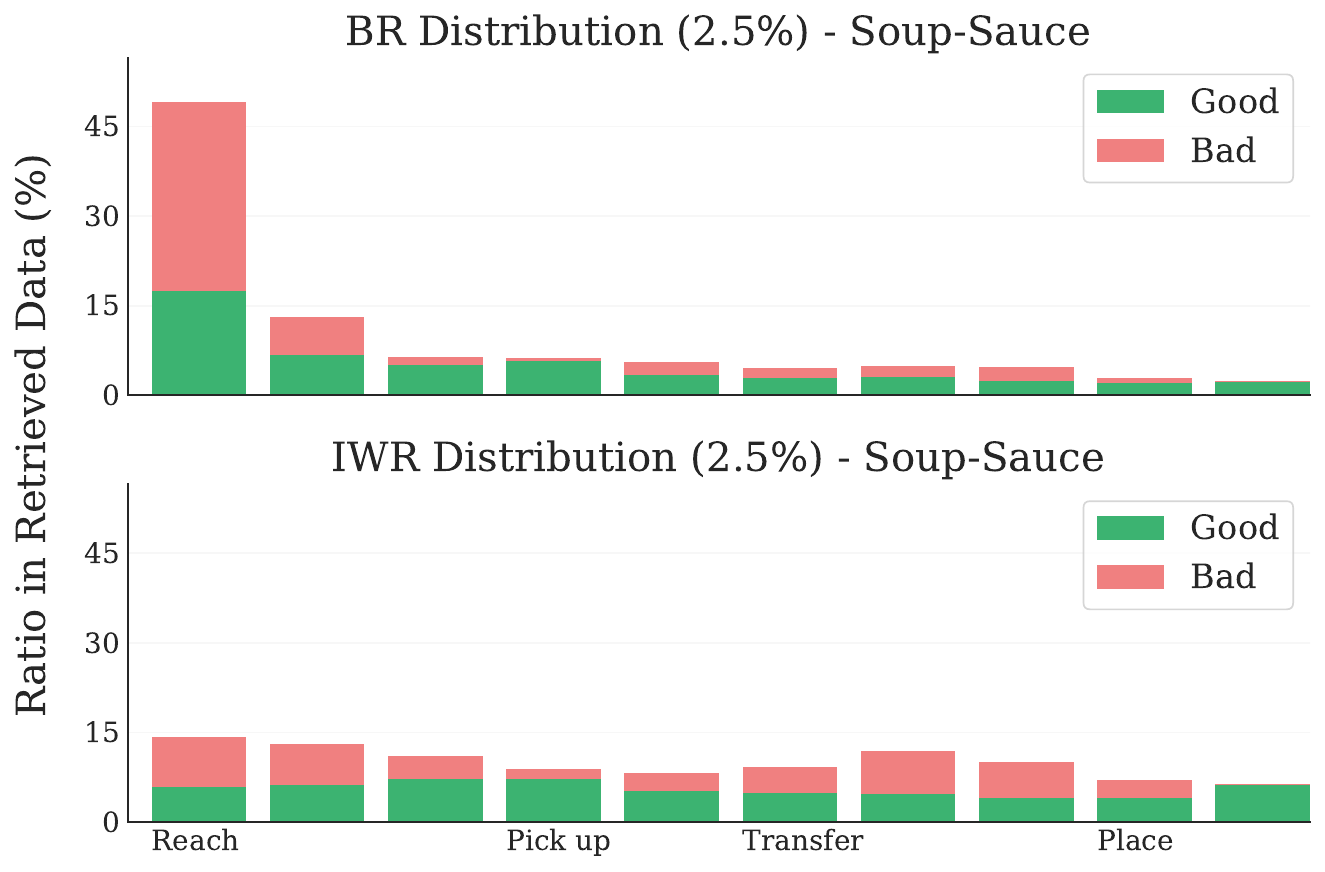}

    \includegraphics[width=\linewidth]{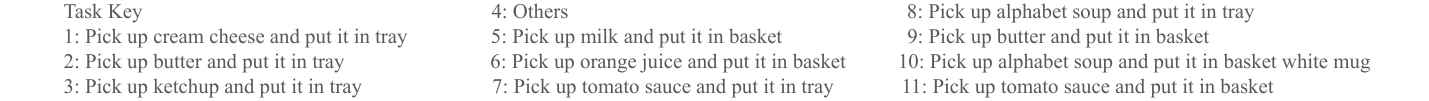}
    \caption{\textbf{Soup-Sauce LIBERO Task.} \textbf{(Left)} Retrieval distribution across tasks. \textbf{(Right)} Retrieval distribution across timesteps.}
    \label{fig:tasks-ret-5}
    \vspace{-0.25in}
\end{figure}

\begin{figure}[h]
    \centering
    \includegraphics[width=\linewidth]{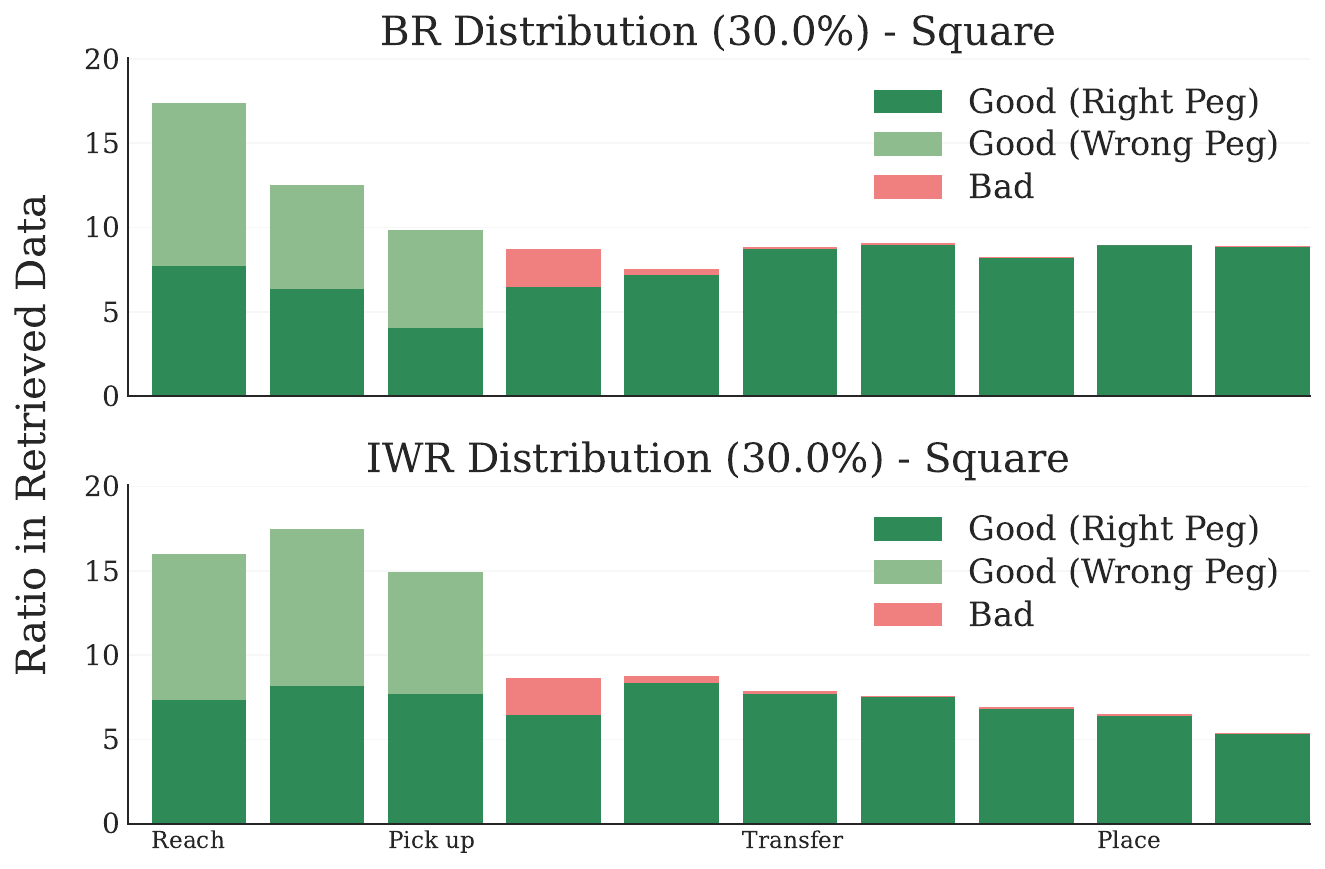}
    \caption{Retrieval distribution across timesteps for \textbf{Robomimic Square Task.}}
    \label{fig:tasks-square-step}
    \vspace{-0.25in}
\end{figure}

In ~\cref{fig:tasks-ret-1} - ~\cref{fig:tasks-ret-5}, we provide additional visualizations of the retrieved data across BR and \abv for all the Libero simulated tasks. In ~\cref{fig:tasks-square-step} we provide visualization for the Robomimic Square task.

\textbf{Retrieval Distribution Across Tasks} For the task-based retrieval visualization (plotted on left in \cref{fig:tasks-ret-1} - ~\cref{fig:tasks-ret-5}), we classify each retrieved task as ``Relevant'' (green), ``Mixed'' (blue) or ``Harmful'' (red). We now explain how we determine whether the task is Relevant, Mixed, or Harmful. First, a task is Relevant if it corresponds exactly to the target task. For example, the target task Mug-Pudding ~\cref{fig:tasks-ret-3} (Put white mug on the plate and put chocolate pudding to right of the plate) has two relevant tasks:  ``Put chocolate pudding to right of the plate'' and ``Put white mug on the plate''. 

We classify tasks from the prior dataset as Mixed if part of the trajectory is similar to the target task. For instance, learning how to pick up a target object is useful, even if the prior task is otherwise different. For instance, for Mug-Pudding, ``Put chocolate pudding to left of plate'' can be useful if we retrieve from the Reach/Pick-up portion of the trajectory, and thus, we label this prior task as Mixed. Similarly, for Mug-Pudding, ``Put red mug on left plate'' is classified as Mixed, because the action of placing an object on the plate is useful for the target task.

The remaining tasks are marked as harmful since they have nothing in common with the target task.

\textbf{Retrieval Distribution Across Timesteps}
For timesteps (plotted on the right in \cref{fig:tasks-ret-1} - ~\cref{fig:tasks-ret-5}), demonstrations are divided into 10 equal bins. 
Green bars represent samples from either relevant tasks or temporally-appropriate portions of partially-relevant mixed tasks (e.g. initial ``Reach'' and ``Pickup'' steps from ``Put chocolate pudding to left of plate'' are relevant to the target task even though the final portion is not). 

Across all plots, we see that \abv consistently helps in both (1) retrieving a higher portion of directly relevant tasks and (2) retrieves a more balanced distribution across timesteps. For Robomimic Square ~\cref{fig:tasks-square-step}, since there are only two prior tasks, both of which are visually very similar, the performance gains in \abv are likely due to retrieving more samples from the reach/pick-up section (grasping the square), which is where most of the failure cases are present.

Note: The visualizations in the Appendix include the following minor changes compared to the figure in the main paper: (1) The legend ``Misc'' is replaced with ``Mixed'' since this labeling better captures the tasks listed under it. (2) The ``Others'' bin is now marked as ``Harmful'' (red) instead of ``Misc'' (blue) since all tasks in the ``Others'' category share neither a common object nor an ending configuration and are therefore adversarial if retrieved. (3) Tasks with similar ending configurations (such as the earlier example,``Put red mug on left plate'') were originally marked as ``Harmful'' instead of ``Misc''/``Mixed'' in the main paper plot, which we have now updated. In order to adhere to these stricter definitions, we have corrected these plots, including an updated version of Figure 4 in the main paper. We plan to update the main paper when possible. Note that these changes do not affect the retrieved values or the conclusions, but instead, more rigorously characterize retrieved tasks.

\end{document}